\documentclass[twocolumn, switch]{article} 

\usepackage{preprint}

\usepackage{enumitem}
\usepackage{amsmath, amsthm, amssymb, amsfonts}
\usepackage{bm}
\usepackage{amsmath}
\usepackage{graphicx}
\usepackage{subfigure}
\usepackage[algoruled,algo2e]{algorithm2e}
\usepackage{tcolorbox}
\usepackage{setspace}
\usepackage{amssymb}
\usepackage{booktabs}
\usepackage{array}
\usepackage{color}
\usepackage{multirow}
\usepackage{multicol}
\usepackage{threeparttable}
\usepackage{url}
\usepackage[page]{appendix}

\usepackage[utf8]{inputenc} 
\usepackage[T1]{fontenc}    
\usepackage{hyperref}       
\usepackage{url}            
\usepackage{booktabs}       
\usepackage{amsfonts}       
\usepackage{nicefrac}       
\usepackage{microtype}      
\usepackage{xcolor}         
\usepackage{graphicx} 
\usepackage{amsmath}
\usepackage{amssymb}
\usepackage{subcaption}
\usepackage{enumitem}
\usepackage{amsthm}

\usepackage{mathtools}

\usepackage{multirow}
\usepackage{colortbl}
\usepackage{xcolor}
\usepackage{amsmath,amssymb,amsthm}
\usepackage{graphicx}
\usepackage{subcaption}
\usepackage{pgffor}

\usepackage{hyperref}
\usepackage{url}
\usepackage{graphicx,wrapfig}
\usepackage{amsmath}
\DeclareMathOperator*{\argmin}{arg\,min}
\usepackage{enumitem}
\usepackage[thicklines]{cancel}
\usepackage{booktabs}
\usepackage{multirow}
\usepackage{algpseudocode}
\usepackage{caption}
\usepackage{wrapfig}

\usepackage{graphics}
\usepackage{hyperref}
\usepackage{color}
\usepackage{multirow}
\usepackage{hhline}
\usepackage{subfigure}
\usepackage{lipsum}

\usepackage[utf8]{inputenc}	
\usepackage[T1]{fontenc}	
\usepackage{xcolor}		

\usepackage{newfloat}
\usepackage{sidecap}
\usepackage{titlesec}

\newtheorem{definition}{Definition}

\definecolor{mygray}{gray}{.9}

\usepackage[numbers,square]{natbib}
\DeclareFloatingEnvironment[name={Supplementary Figure}]{suppfigure}

\sidecaptionvpos{figure}{c}

\titlespacing\section{0pt}{12pt plus 3pt minus 3pt}{1pt plus 1pt minus 1pt}
\titlespacing\subsection{0pt}{10pt plus 3pt minus 3pt}{1pt plus 1pt minus 1pt}
\titlespacing\subsubsection{0pt}{8pt plus 3pt minus 3pt}{1pt plus 1pt minus 1pt}

\newcommand\blfootnote[1]{%
\begingroup
\renewcommand\thefootnote{}\footnote{#1}%
\addtocounter{footnote}{-1}%
\endgroup
}

\title{Training-Free Reasoning and Reflection in MLLMs}

\usepackage{authblk}

\author[1]{Hongchen Wei}
\author[1]{Zhenzhong Chen*}
\affil[1]{School of Remote Sensing and Information Engineering, Wuhan University}

\begin{document}
\twocolumn[ 
  \begin{@twocolumnfalse} 
  
\maketitle

\begin{abstract}
    Recent advances in Reasoning LLMs (e.g., DeepSeek-R1 and OpenAI-o1) have showcased impressive reasoning capabilities via reinforcement learning. 
    However, extending these capabilities to Multimodal LLMs (MLLMs) is hampered by the prohibitive costs of retraining and the scarcity of high-quality, verifiable multimodal reasoning datasets. 
    This paper introduces \textbf{FRANK} Model, a training-\textbf{FR}ee \textbf{AN}d r1-li\textbf{K}e MLLM that imbues off-the-shelf MLLMs with reasoning and reflection abilities, without any gradient updates or extra supervision. 
    Our key insight is to decouple perception and reasoning across MLLM decoder layers. 
    Specifically, we observe that compared to the deeper decoder layers, the shallow decoder layers allocate more attention to visual tokens, while the deeper decoder layers concentrate on textual semantics. 
    This observation motivates a hierarchical weight merging approach that combines a visual-pretrained MLLM with a reasoning-specialized LLM. 
    To this end, we propose a layer-wise, Taylor-derived closed-form fusion mechanism that integrates reasoning capacity into deep decoder layers while preserving visual grounding in shallow decoder layers. 
    Extensive experiments on challenging multimodal reasoning benchmarks demonstrate the effectiveness of our approach. 
    On the MMMU benchmark, our model FRANK-38B achieves an accuracy of 69.2, outperforming the strongest baseline InternVL2.5-38B by +5.3, and even surpasses the proprietary GPT-4o model. 
    Our project homepage is at: \url{http://iip.whu.edu.cn/frank/index.html}
  \end{abstract}

\vspace{0.4cm}

  \end{@twocolumnfalse} 
]

\section{INTRODUCTION}
\label{sec:intro}
\blfootnote{Corresponding author: Zhenzhong Chen, E-mail:zzchen@ieee.org}
Recent reasoning-focused large language models (LLMs)~\cite{guo2025deepseek,team2025kimi,ma2025s,LightmanKBEBLLS24} such as DeepSeek-R1~\cite{guo2025deepseek} and OpenAI-o1~\cite{OpenAI} have demonstrated strong performance in tasks requiring complex logic, including math reasoning, symbolic manipulation, and program synthesis. 
  These models leverage mechanisms like reinforcement learning to perform multi-step problem-solving and iterative self-correction, often surpassing even human experts. 
  
  In real-world scenarios, numerous tasks demand sophisticated multimodal reasoning capabilities. 
  For instance, solving visual mathematics problems, interpreting diagrams, and understanding code snippets embedded within images require the integration of visual perception with logical reasoning. 
  Inspired by the successes of reasoning-augmented LLMs, researchers have begun exploring methods to endow multimodal large language models (MLLMs)~\cite{xu2024llava,dong2024insight,yao2024mulberry,liu2025visual} with similar reasoning abilities. 
  A prevalent approach~\cite{huang2025vision,peng2025lmm,zhang2025r1} involves adapting reinforcement learning techniques, such as GRPO~\cite{guo2025deepseek}, to the multimodal context. 
  However, this strategy encounters significant challenges. 
  Firstly, the reinforcement learning training of large-scale MLLMs demands substantial computational resources, making it resource-intensive. 
  Secondly, there is a notable scarcity of high-quality, verifiable multimodal reasoning datasets, which are essential for effective training. 
  This paucity of suitable data severely impedes the development and scalability of reasoning-capable MLLMs.
  \begin{figure*}[tbp]
    \centering
      \includegraphics[width=0.95\textwidth]{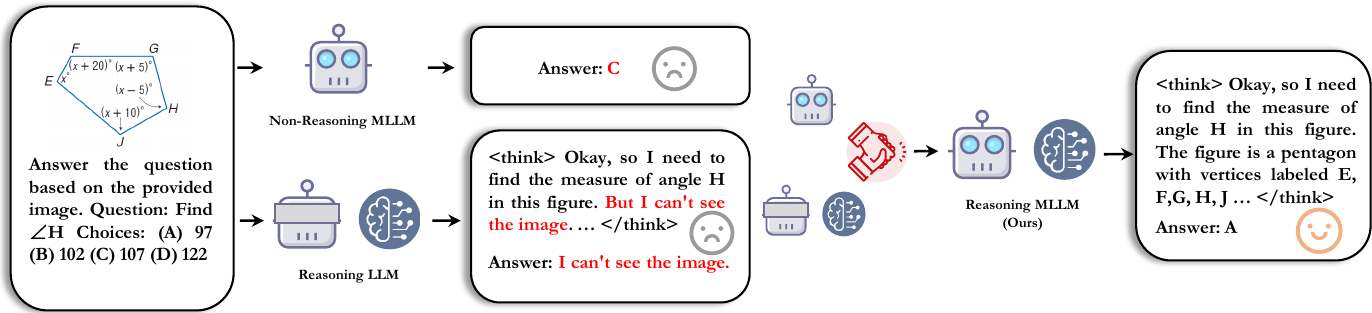}
    \caption{Non-reasoning MLLMs lack reasoning and reflection abilities, while reasoning LLMs are unable to perceive visual information. 
    We propose a training-free, closed-form layerwise fusion method that combines visual perception and language reasoning strengths, substantially enhancing overall reasoning capability in multimodal settings.}
    \label{fig:vis}
  \end{figure*}
  
  In this paper, we introduce \textbf{FRANK}, a training-free and R1-like MLLM that is designed to endow existing MLLMs with advanced reasoning and reflection capabilities without any additional training or supervision. 
  Figure~\ref{fig:vis} visualizes this pipeline. 
  Our method is built upon two key insights: 
  \begin{itemize}[leftmargin=*, noitemsep, topsep=0pt]
    \item \textbf{Homologous Model Merging}: 
    We conceptualize MLLMs as base LLMs fine-tuned on visual-text data, while reasoning-specialized LLMs represent the same base LLM fine-tuned on reasoning tasks. 
    According to the task arithmetic hypothesis~\cite{WortsmanIGRLMNF22,IlharcoRWSHF23,ZhangCLH23}, the difference in weights between a fine-tuned model and its base model encapsulates the task-specific adaptations. 
    By merging task vectors from models fine-tuned on different tasks, we can integrate multiple capabilities into a single model without additional training. 
    \item \textbf{Layer-wise Functional Specialization in MLLMs}: 
    Drawing inspiration from the hierarchical processing observed in the human brain~\cite{brincat2018gradual,kawasaki2022hierarchical}, where sensory inputs are initially processed in primary sensory areas and progressively integrated into higher-order cognitive functions in association cortices, we observe a similar pattern in MLLMs. 
    As shown in Figure~\ref{fig:prior}, compared to the deeper decoder layers, the shallow decoder layers allocate more attention to visual tokens, facilitating perceptual grounding, while the deeper decoder layers concentrate on textual semantics. 
  \end{itemize}
  
  Based on these two key insights, we design a hierarchical weight merging strategy to effectively integrate a vision-text pretrained MLLM with a reasoning-specialized LLM. 
  Specifically, we develop a layer-wise, Taylor-derived closed-form fusion mechanism that enables fine-grained control over the contribution of each model at different depths of the decoder. 
  This mechanism builds on the task vector formulation and refines it with a layer-wise optimization strategy: For each decoder block, we derive a closed-form solution for the optimal task vector fusion weights by minimizing the Taylor-approximated task loss difference. 
  This allows precise control over how visual and reasoning adaptations are combined at each layer. Furthermore, guided by the empirical prior that shallow decoder layers in MLLMs attend more to visual inputs while deeper layers focus on symbolic reasoning, we incorporate layer-dependent fusion weights to align with the distinct functional roles across the model hierarchy. 
  This design enables FRANK to embed reasoning capabilities into deeper layers, responsible for abstraction and reasoning, while preserving the visual grounding in shallower layers, which are more sensitive to perceptual signals. 
  
  We validate the effectiveness of FRANK through comprehensive evaluations on challenging multimodal reasoning benchmarks. 
  Notably, our model FRANK-38B achieves an accuracy of 69.2 on the MMMU benchmark, outperforming the strongest baseline InternVL2.5-38B by +5.3, and even surpasses the proprietary GPT-4o model. 
  These results underscore FRANK's ability to enhance reasoning capabilities in MLLMs without additional training or supervision.
  
  In summary, our contributions are threefold:
  \begin{itemize}[topsep=0pt]
    \item The layer-wise functional specialization in MLLMs, where shallow decoder layers focus on visual perception and deeper layers on textual reasoning, is identified and leveraged.
    \item A novel, training-free hierarchical weight fusion mechanism is proposed, integrating reasoning capabilities into existing MLLMs by merging task vectors at each layer, guided by a Taylor-derived closed-form solution.
    \item It is demonstrated that FRANK effectively enhances the reasoning and reflection abilities of MLLMs, achieving superior performance on multimodal reasoning tasks without the need for additional training or supervision.
  \end{itemize}
  
  \section{Related Work}
  \subsection{Multimodal Large Language Models}
  Multimodal large language models (MLLMs) have recently advanced the integration of visual and linguistic understanding through large-scale vision-language pretraining and instruction tuning. 
  Early works~\cite{VisionLLM,bai2023qwen,LiuLWL23a,SunYCZZWGL0W24} enable zero- or few-shot visual question answering, image captioning, and multimodal instruction following by augmenting an LLM with a pretrained visual encoder. 
  While these models excel at aligning visual features with text, their performance on tasks requiring multi-step logical reasoning or self-reflection remains limited. 
  
  Building on chain-of-thought (CoT)~\cite{Wei0SBIXCLZ22} prompting in pure-text LLMs, several recent works~\cite{xu2024llava,dong2024insight,yao2024mulberry,liu2025visual,huang2025vision,peng2025lmm,zhang2025r1} attempt to endow MLLMs with explicit reasoning capabilities. 
  LLaVA-Reasoner~\cite{zhang2024improve} shows that naively appending CoT prompts often yields marginal gains or can even degrade MLLM accuracy. 
  To better guide intermediate reasoning, CCoT~\cite{MitraHDH24} and TextCoT~\cite{luan2024textcot} introduce plan-based CoT prompting, prompting the model to generate structured outlines before deriving final answers. 
 In concurrent work, methods like MM-Eureka~\cite{meng2025mm}, Vision-R1~\cite{huang2025vision}, and LMM-R1~\cite{peng2025lmm} extend reinforcement-learning-based reasoning (e.g., GRPO) to the multimodal setting, using multimodal datasets to fine-tune MLLMs. 
  However, all of these approaches require costly secondary training and depend on the availability of large, high-quality multimodal reasoning corpora, which remain extremely scarce, thus limiting their scalability and generalization.
  
  \subsection{Model Merging}
  Model merging~\cite{MatenaR22,WortsmanIGRLMNF22,IlharcoRWSHF23,YadavTCRB23,ZhouSWC24,Jin0P023} has emerged as an efficient alternative to multi-task fine-tuning, enabling the fusion of multiple task-specific models into a single model without accessing original training data. 
  Early work on weight averaging~\cite{WortsmanIGRLMNF22} showed that simple parameter interpolation can improve robustness across tasks. 
  Task Arithmetic~\cite{IlharcoRWSHF23} generalized this idea by introducing task vectors that represent the differences between fine-tuned and base model weights, and linearly combining them to incorporate new capabilities. 
  To address interference between tasks, TIES-Merging ~\cite{YadavTCRB23} trims low-magnitude updates and aligns parameter signs before merging, while DARE~\cite{Yu0Y0L24} further sparsifies and rescales delta parameters to reduce redundancy. 
  MetaGPT~\cite{ZhouSWC24} formalizes model merging as a multi-task learning problem, achieving balanced task performance by computing the norms of different model parameters. 
  Although these methods achieve strong results for text-only LLMs, they do not exploit the unique structure of multimodal models. 
  In concurrent work, VLM-Merging~\cite{chen2025bring} employs fixed fusion weights (with an MLLM weight of 0.9 and an LLM weight of 0.1) to merge MLLM and LLM, enabling the MLLM to acquire reasoning capabilities. 
  However, such manually designed fusion weights are often suboptimal and struggle to transfer the LLM's reflective abilities effectively. 
  In this work, we extend the model merging paradigm to MLLMs by incorporating multimodal priors, specifically, layer-wise perception-to-cognition specialization, into the fusion process, thereby enabling training-free integration of visual perception and reasoning capabilities.  
  
  \section{Approach}
  In this section, we introduce FRANK, a training-free and R1-like MLLM that endows off-the-shelf MLLMs with advanced reasoning and reflection by merging them with reasoning-specialized LLMs, entirely without gradient updates or extra supervision. 
  At its core, FRANK relies on two principles: 
  (1) \textbf{homologous model merging}, which treats both models as task-fine-tuned variants of the same base LLM and fuses them via task-vector arithmetic; 
  and (2) \textbf{hierarchical layer-wise fusion}, which exploits the observation that shallow decoder layers chiefly process visual inputs while deep layers focus on text, allowing us to inject reasoning only where it is most effective.
  
  \subsection{Preliminary: Task Arithmetic for Homologous Model Merging}
  We build on the paradigm of task arithmetic~\cite{IlharcoRWSHF23}, which provides a simple yet effective mechanism to merge multiple fine-tuned models, so-called homologous models, that share the same base architecture. 
  Let $f(\cdot\,;\theta_0)$ denote a pre-trained base model, and let  
  \begin{equation}
    \theta_t = \arg\min_{\theta} \mathcal{L}_t\bigl(f(\cdot\,;\theta),\mathcal{D}_t\bigr)
    \label{eq:mllm}
  \end{equation}
  be the parameters obtained by fine-tuning on task~$ t $ with loss $ \mathcal{L}_t $ over dataset $ \mathcal{D}_t $. We define the task vector for task~$ t $ as $\tau_t \;=\;\theta_t - \theta_0\,.$ 
  Under the homologous assumption that all $ \theta_t $ lie in the same parameter space as $ \theta_0 $, we can form a merged model by a linear combination of these task vectors:  
  \begin{equation}
    \theta_{f}
    = \theta_0 \;+\;\sum_{t \in \{V, R\}} \lambda_t\,\tau_t\;,
    \label{eq:task_arithmetic}
  \end{equation}
  where $ \{\lambda_t\} $ are non-negative fusion weights controlling each task's contribution.  
  
  Early approaches choose all $ \lambda_t $ heuristically (e.g., a constant $ 0.3 $) or via grid search on held-out data, but these methods either under-utilize model capacity or incur prohibitive search costs as $ T $ grows. 
  
  \subsection{FRANK: a Training-free and R1-like MLLM}
  
  Building on the task-arithmetic preliminary, FRANK introduces two key innovations: 
  \textbf{Layer-wise Fusion}, which respects the functional specialization of each decoder layer (for clarity, we define a decoder's block as a ``layer''); 
  \textbf{Modality Priors}, which steer shallow layers toward visual grounding and deep layers toward symbolic reasoning.
  We next detail (i) how we decompose the MLLM decoder into per-layer task vectors; (ii) the Taylor-based derivation of closed-form fusion weights at each layer; and (iii) the incorporation of layer-dependent modality priors.
  
  \subsubsection{Layer-Wise Fusion Setup for MLLMs}
\begin{figure}[tbp]
    \centering
      \includegraphics[width=0.45\textwidth]{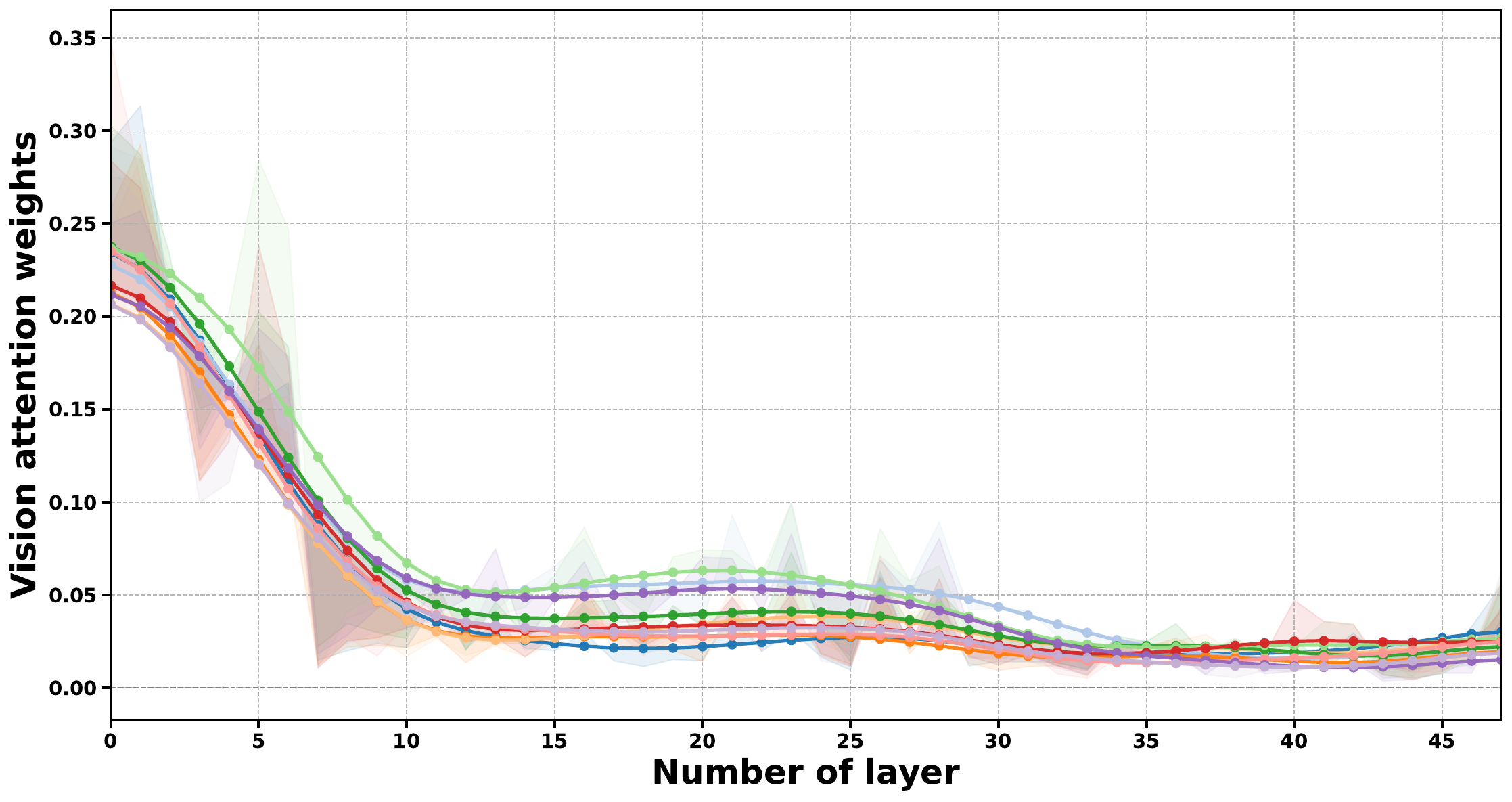}
      \caption{Layer-wise visual attention of NVIL-15B. 
    Each curve shows the average attention from a text token to all visual tokens across layers. 
    Shallow layers assign significantly higher attention to visual tokens, while attention in deeper layers approaches zero and rapidly descends indicating a shift from perception to language reasoning. 
    This supports our use of an exponential decay prior to the fusion process. }
      \label{fig:prior}
    \end{figure}

  FRANK bridges these two paradigms by viewing the vision-fine-tuned decoder and a reasoning-specialized decoder as ``homologous'' variants of the same base model checkpoint. Leveraging their shared architecture, we can merge their strengths without retraining.
  The decoder consists of $L$ stacked transformer layers, indexed by $l=1,\dots,L$. 
  Figure~\ref{fig:prior} shows that, compared to the deeper decoder layers, the shallow decoder layers allocate more attention to visual tokens, facilitating perceptual grounding, while the deeper decoder layers concentrate on textual semantics. 
  To preserve this functional hierarchy and prevent interference between vision and reasoning, FRANK performs layer-wise weight fusion: each decoder layer is merged independently, preserving its specialized role.
  
  Concretely, let $
    \theta_0^{(l)}, 
    \theta_V^{(l)}, 
    \theta_R^{(l)} 
  $ denote the parameters of layer $l$ for the pre-trained base model, the vision-fine-tuned MLLM, and the reasoning-fine-tuned LLM, respectively. We define the layer-wise task vectors: 
  \begin{equation}
    \tau_V^{(l)} = \theta_V^{(l)} - \theta_0^{(l)}, \qquad
    \tau_R^{(l)} = \theta_R^{(l)} - \theta_0^{(l)}.
  \end{equation}
  These differences capture how each task (vision vs.\ reasoning) shifts the model weights at each depth.
  
  To measure the impact of fusing these shifts, we introduce two metrics. 
  First, the Layer-Wise Task Loss Difference (LTLD) compares the fused layer's loss against each branch's own fine-tuned layer:
  
  \begin{definition}[Layer-Wise Task Loss Difference, LTLD]
  Let $\mathcal{L}_t^{(l)}(\theta,x)$ be the loss of branch $t\in\{V,R\}$ at layer $l$ on input $x$. For fusion weights $(\lambda_V,\lambda_R)$, define the fused parameter
  \begin{equation}
    \theta_f^{(l)} = \theta_0^{(l)} + \lambda_V\,\tau_V^{(l)} + \lambda_R\,\tau_R^{(l)}, 
  \end{equation}
  where, $\theta_f^{(l)}$ represents the parameters of the fusion model at layer $l$, while $\tau_V^{(l)}$ and $\tau_R^{(l)}$ denote the task vectors at layer $l$ of the non-reasoning MLLM and reasoning-specialized LLM, respectively. 
  Then, we define the layer-wise task loss difference (LTLD) as
  \begin{equation}
    \mathrm{LTLD}_t^{(l)}(\lambda_V,\lambda_R)
    = \mathcal{L}_t^{(l)}(\theta_f^{(l)},x) - \mathcal{L}_t^{(l)}(\theta_b^{(l)},x),
  \end{equation}
  quantifies the degradation when using the fused weights in place of branch $b$'s own.
  \end{definition}
  
  \begin{definition}[Layer-Wise Average Loss Difference, LALD]
  The LALD averages LTLD across both branches, 
  \begin{equation}
    \mathrm{LALD}^{(l)}(\lambda_V,\lambda_R)
    = \tfrac12\bigl(\mathrm{LTLD}_V^{(l)} + \mathrm{LTLD}_R^{(l)}\bigr),
  \end{equation}
  and serves as our per-layer fusion objective. 
  By minimizing LALD independently at each layer, we derive fusion weights that optimally balance visual grounding and reasoning without cross-layer interference. 
  \end{definition}
  
  We emphasize that LTLD is a theoretical sensitivity measure, not an operation we perform in implementation. 
  Specifically, LTLD represents the second-order Taylor approximation of the loss increase induced by small perturbations in the parameters of layer $l$, holding all other layers fixed. 
  Under the Neural Tangent Kernel (NTK)~\cite{JacotHG18} linearization and task-vector orthogonality assumptions, this localized view yields a tractable quadratic form in the norms $|\tau_V^{(l)}|^2$ and $|\tau_R^{(l)}|^2$, from which we derive closed-form fusion weights (see Section~\ref{sec:taylor_approx}). 
  Importantly, we do not compute per-layer losses on the fly; LTLD merely guides our analytical derivation. 
  In subsequent sections, we will show how to approximate LTLD via a second-order Taylor expansion, invoke NTK linearization and vector orthogonality, and obtain a data-agnostic closed-form for $(\lambda_V^{(l)},\lambda_R^{(l)})$.

\subsubsection{Taylor-Based Approximation and Closed-Form Fusion Weights}
\label{sec:taylor_approx}
To obtain efficient, data-agnostic fusion weights at each decoder layer, we approximate the layer-wise average loss difference (LALD) using a second-order Taylor expansion. Intuitively, this expansion captures how a small perturbation---arising from merging vision and reasoning task vectors---affects the loss. 
Under two standard assumptions: 

\textbf{NTK Linearization.} 
In the infinite-width regime, neural networks evolve under training according to a fixed NTK~\cite{JacotHG18}, which implies that small weight perturbations produce locally linear changes in the model output~\cite{JacotHG18,ZhouSWC24}. 
The appendix~\ref{app:proof-NTK} provides additional details. 
Previous study~\cite{ZhouSWC24} has empirically validated this NTK-linear behavior for LLMs. 
They evaluated LLaMA-2-7b-chat-hf~\cite{touvron2023llama-2} on the AGIEval benchmark~\cite{ZhongCGLLWSCD24}, sampling three random prompts and measuring model outputs for interpolation coefficients $\alpha\in\{0,0.1,\dots,1.0\}$. 
The output trajectories scale almost perfectly linearly with $\alpha$, confirming that LLM operates in the NTK regime during fine-tuning, which is specifically suitable for the LLM's arithmetic scenario. 

\begin{figure}[tbp]
      \centering
      \includegraphics[width=0.45\textwidth]{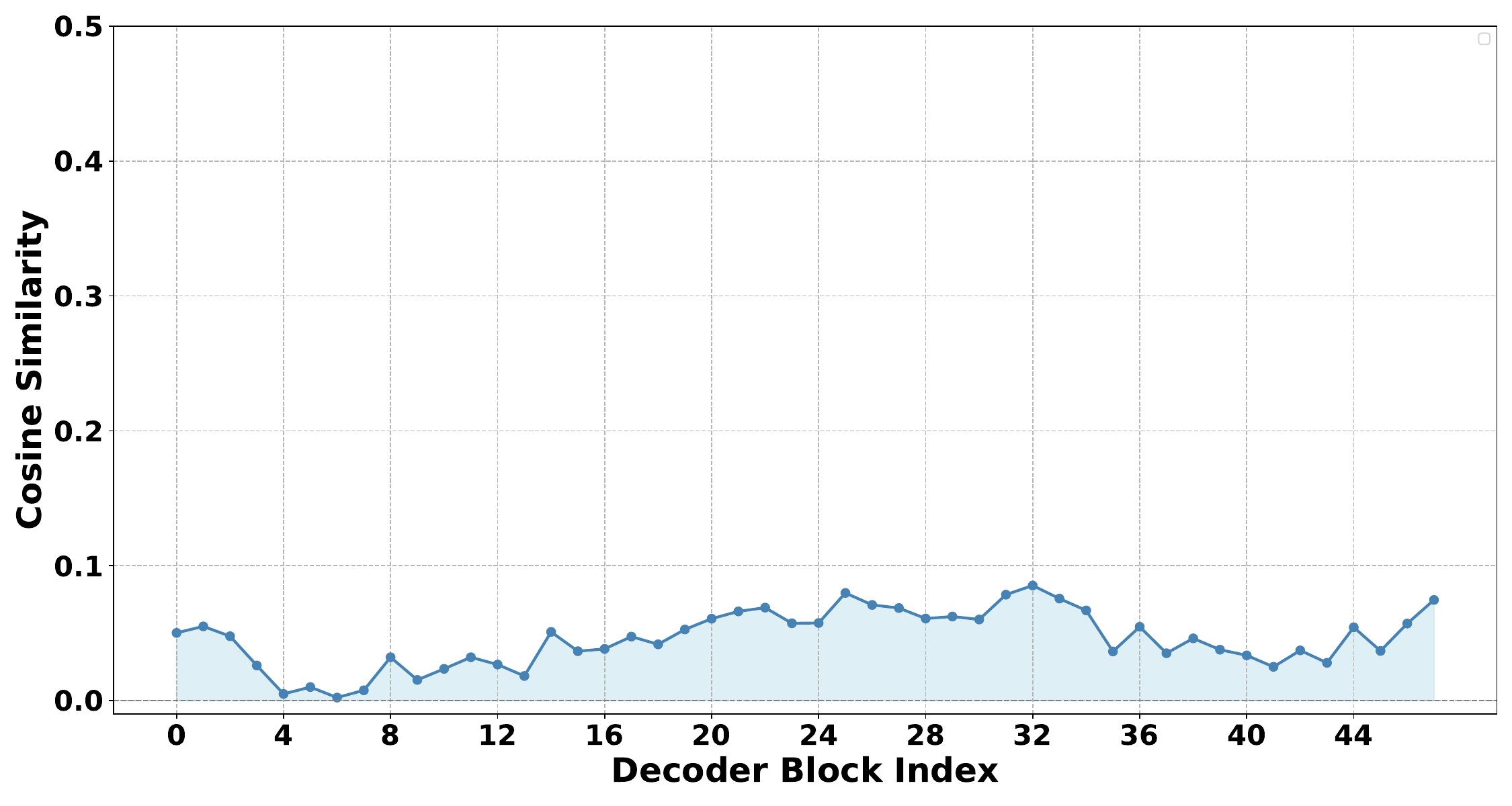}
      \caption{Cosine similarity between task vectors of vision-finetuned (NVIL-15B) and reasoning-finetuned (DeepSeekDistil-Qwen2.5-14B) models at each decoder block. The task vector at each block is computed by flattening the weight deltas with respect to the base model. The similarity remains close to 0 across all layers, indicating strong near-orthogonality.}
      \label{fig:cosine}
\end{figure}

\textbf{Task-Vector Orthogonality.} 
Although both vision and reasoning fine-tuning update the same decoder weights, their resulting task vectors often lie in nearly orthogonal subspaces. 
As shown in Figure~\ref{fig:cosine}, we verify this by computing the cosine similarity between $\tau_V^{(l)}$ and $\tau_R^{(l)}$ at each layer, which remains close to zero across all layers. 
The appendix~\ref{app:proof-orthogonality} provides additional details.  

Under the NTK linearization and task-vector orthogonality assumptions, we can now quantify how merging vision and reasoning updates perturbs the layer loss.  
In the next step, we expand each task's loss around its fine-tuned parameters to derive a closed-form bound on the loss increase. 

Let $\theta_0^{(l)}$ be the initialization at layer~$l$, and $\theta_t^{(l)}=\theta_0^{(l)}+\tau_t^{(l)}$ be the fine-tuned weights for task $t\in\{V,R\}$.  Define the fused weights
\begin{equation}
\theta_f^{(l)} = \theta_0^{(l)} + \sum_{t\in\{V,R\}}\lambda_t^{(l)}\,\tau_t^{(l)}
\end{equation}
and the fusion residual
\begin{equation}
h_t^{(l)} = \theta_f^{(l)} - \theta_t^{(l)}
    = \sum_{k\neq t \in \{V, R\}}\lambda_k^{(l)}\,\tau_k^{(l)} - (1-\lambda_t^{(l)})\,\tau_t^{(l)}.
\end{equation}
Around $\theta_t^{(l)}$, a second-order Taylor expansion of the layer loss $\mathcal{L}_t^{(l)}$ gives
\begin{equation}
\mathcal{L}_t^{(l)}(\theta_f^{(l)}) \,\approx\,
    \mathcal{L}_t^{(l)}(\theta_t^{(l)})
    + \underbrace{\nabla\mathcal{L}_t^{(l)}(\theta_t^{(l)})^\top h_t^{(l)}}_{ \approx0 } 
    + \tfrac12\,h_t^{(l)\top}\nabla^2\mathcal{L}_t^{(l)}(\theta_t^{(l)})\,h_t^{(l)},
\end{equation}
where the first-order term vanishes under near-convergence, as guaranteed by the NTK theory for wide networks.
This NTK regime further implies two key properties for the second-order term: 
1) The Hessian is dominated by the Jacobian Gram matrix due to the approximate linearity of the network in parameter space, and 
2) Its eigenvalue distribution becomes approximately isotropic. 
Concretely, the Hessian admits the following approximation:
\begin{equation}
\nabla^2\mathcal{L}_t^{(l)}(\theta_t^{(l)})
    \approx \delta_t^{(l)}\,I_{d_l},
    \quad
    \delta_t^{(l)} = \tfrac{1}{d_l}\mathrm{tr}\Bigl(\nabla f^{(l)}(x_t;\theta_0^{(l)})\nabla f^{(l)}(x_t;\theta_0^{(l)})^\top\Bigr), 
\end{equation}
where, $\delta_t^{(l)}$ is a data-dependent constant, $d_l$ represents the parameter dimension of the $l$-th layer, $I_{d_l}$ denotes the $d_l$-dimensional identity matrix, which is used to simplify the structure of the Hessian. 
Consequently,
\begin{equation}
\mathrm{LTLD}_t^{(l)}
    = \mathcal{L}_t^{(l)}(\theta_f^{(l)}) - \mathcal{L}_t^{(l)}(\theta_t^{(l)})
    \le \tfrac{\delta_t^{(l)}}{2}\,\|h_t^{(l)}\|^2.
\end{equation}
Summing the above bound for $t \in \{V,R\}$ and using $\langle\tau_V^{(l)},\tau_R^{(l)}\rangle\approx0$ yields a layer-wise bound on the average loss increase:
\begin{equation}
\mathrm{LALD}^{(l)}
    \le \tfrac12 \sum_{t\in\{V,R\}}\delta_t^{(l)}\,
    \Bigl[(1-\lambda_t^{(l)})^2\|\tau_t^{(l)}\|^2
    + \sum_{k\neq t \in\{V,R\}}(\lambda_k^{(l)})^2\|\tau_k^{(l)}\|^2\Bigr]\,\|\tau_t^{(l)}\|^2.
\end{equation}
To derive the optimal fusion weights, we minimize the LALD bound under the constraint $\lambda_V+\lambda_R=1$. 
Setting the derivative to zero and solving yields the closed-form solution: 
\begin{equation}
    \lambda_t^{(l)} = \frac{\|\tau_t^{(l)}\|^2}{\|\tau_V^{(l)}\|^2 + \|\tau_R^{(l)}\|^2},
    \quad t\in\{V,R\}.
\end{equation}

This compact expression emphasizes FRANK's key advantage: fusion weights depend solely on the observed parameter shifts of each task, eliminating the need for held-out data, grid search, or additional training. 
For full statements of the Taylor expansion lemma, formal properties, and proofs leading to this closed form, please refer to Appendix~\ref{app:proof-ltld} and ~\ref{app:proof-weights}.
    \begin{table*}[t]
    \captionsetup{font={footnotesize}}
    \caption{Comparison of FRANK variants (8B, 15B, 38B) and state-of-the-art baselines across five multimodal reasoning benchmarks: MMMU val, MMMU-Pro standard (10 opts), MathVista testmini, MathVision testmini, and WeMath testmini. 
    * indicates the baseline model.}
    \centering
    \footnotesize
    \renewcommand{\arraystretch}{1.1} 
    \setlength{\tabcolsep}{1.9pt} 
    \begin{tabular}{m{0.5cm}|l|c|cccccc}
    \toprule[1.6pt]
    & {\footnotesize \textbf{Methods}} & \textit{\footnotesize \textbf{Size}} & {\footnotesize \textbf{MMMU}} & {\footnotesize \textbf{MMMU-Pro}} & {\footnotesize \textbf{MathVista}} & {\footnotesize {\textbf{MathVision}}} & {\footnotesize \textbf{WeMath}} \\ \midrule[0.7pt]
     \multirow{7}{*}[1ex]{\rotatebox{90}{\normalsize \small{\textbf{Samll}}}} 
    & {\small LLaVA-1.5~\cite{LiuLLL24}}${}_{\textcolor{red}{\rm ~[\text{2023.10.05}]}}$ & 7B & {\small 35.7} & {\small 19.7} & {\small 25.6} & {\small 10.2} & {\small 7.0} \\
    & {\small LLaVA-NeXT~\cite{LLaVA-NeXT}}${}_{\textcolor{red}{\rm ~[\text{2024.01.30}]}}$ & 7B & {\small 35.3} & {\small 19.4} & {\small 24.9} & {\small 10.0} & {\small 3.3} \\
    & {\small LLaVA-LLaMA3~\cite{LLaVA-NeXT}}${}_{\textcolor{red}{\rm ~[\text{2024.05.10}]}}$ & 8B & {\small 39.2} & {\small -} & {\small 40.0} & {\small -} & {\small -} \\
    & {\small VILA1.5-LLaMA3~\cite{LinYP0SH24}}${}_{\textcolor{red}{\rm ~[\text{2024.05.16}]}}$ & 8B & {\small 38.6} & {\small -} & {\small 36.7} & {\small -} & {\small -} & {\small  } \\
    \cmidrule(r){2-9}
    & {\small Idefics3-LLaMA3*~\cite{laurenccon2024building}}${}_{\textcolor{red}{\rm ~[\text{2024.08.07}]}}$ & 8B & {\small 43.9} & {\small 32.6} & \textbf{\small 58.4} & {\small 20.1} & {\small \textbf{12.3}} \\
    & \cellcolor{gray!25}{\small \textbf{FRANK}~(Ours)} & \cellcolor{gray!25}8B & \cellcolor{gray!25}{\small \textbf{48.3}} & \cellcolor{gray!25}{\small \textbf{34.7}} & \cellcolor{gray!25}{\small 50.7} & \cellcolor{gray!25}{\small \textbf{27.6}} & \cellcolor{gray!25}{\small {11.7}} \\
    \midrule[0.7pt]
    \midrule[0.7pt]
     \multirow{7}{*}[1ex]{\rotatebox{90}{\normalsize \footnotesize{\textbf{Medium}}}} 
    & {\small LLaVA-1.5~\cite{LiuLLL24}}${}_{\textcolor{red}{\rm ~[\text{2023.10.05}]}}$ & 13B & {\small 37.0} & {\small -} & {\small 27.7} & {\small 13.1} & {\small 7.4} \\
    & {\small ShareGPT4V~\cite{ChenLDZHWZL24}}${}_{\textcolor{red}{\rm ~[\text{2023.11.21}]}}$ & 13B & {\small 36.6} & {\small -} & {\small 29.3} & {\small 13.9} & {\small -} \\
    & {\small LLaVA-NeXT~\cite{LLaVA-NeXT}}${}_{\textcolor{red}{\rm ~[\text{2024.01.30}]}}$ & 13B & {\small 36.2} & {\small 19.8} & {\small -} & {\small -} & {\small -} \\
    & {\small VILA-1.5~\cite{LinYP0SH24}}${}_{\textcolor{red}{\rm ~[\text{2024.05.16}]}}$ & 13B & {\small 37.9} & {\small -} & {\small 42.7} & {\small 15.2} & {\small 11.4} \\
    \cmidrule(r){2-9}
    & {\small NVILA*~\cite{liu2024nvila}}${}_{\textcolor{red}{\rm ~[\text{2024.12.05}]}}$ & 15B & {\small 53.2} & {\small 36.2} & {\small \textbf{67.6}} & {\small 23.2} & {\small 31.1} \\
    & \cellcolor{gray!25}{\small \textbf{FRANK}~(Ours)} & \cellcolor{gray!25}15B & \cellcolor{gray!25}{\small \textbf{61.3}} & \cellcolor{gray!25}{\small \textbf{49.4}} & \cellcolor{gray!25}{\small {55.4}} & \cellcolor{gray!25}{\small \textbf{37.2}} & \cellcolor{gray!25}{\small \textbf{32.3}} \\
    \midrule[0.7pt]
    \midrule[0.7pt]
     \multirow{9}{*}[1ex]{\rotatebox{90}{\normalsize \textbf{Large}}} 
    & {\small LLaVA-NeXT~\cite{LLaVA-NeXT}}${}_{\textcolor{red}{\rm ~[\text{2024.01.30}]}}$ & 34B & {\small 48.1} & {\small 30.3} & {\small 46.5} & {\small -} & {\small -} \\
    & {\small VILA-1.5~\cite{LinYP0SH24}}${}_{\textcolor{red}{\rm ~[\text{2024.05.16}]}}$ & 40B & {\small 51.9} & {\small 35.9} & {\small 49.5} & {\small -} & {\small -} \\
    & {\small LLaVA-OneVision~\cite{li2024llava}}${}_{\textcolor{red}{\rm ~[\text{2024.08.06}]}}$ & 72B& {\small 56.8} & {\small 38.0} & {\small 67.5} & {\small 25.3} & {\small 32.0} \\
    & {\small Qwen2-VL~\cite{wang2024qwen2}}${}_{\textcolor{red}{\rm ~[\text{2024.09.18}]}}$ & 72B & {\small 64.5} & {\small 49.2} & {\small 70.5} & {\small 26.6} & {\small 36.0} \\
    & {\small GPT-4o~\cite{GPT-4o}}${}_{\textcolor{red}{\rm ~[\text{2024.05.13}]}}$ & - & {\small 69.1} & {\small 54.0} & {\small 63.8} & {\small 29.9} & {\small -} \\
    \cmidrule(r){2-9}
    & {\small InternVL2.5*~\cite{chen2024expanding}}${}_{\textcolor{red}{\rm ~[\text{2024.12.06}]}}$ & 38B & {\small 63.9} & {\small 48.0} & {\small 71.9} & {\small 32.2} & {\small 38.3} \\
    & \cellcolor{gray!25}{\small \textbf{FRANK}~(Ours)} & \cellcolor{gray!25}38B & \cellcolor{gray!25}{\small \textbf{69.2}} & \cellcolor{gray!25}{\small \textbf{56.8}} & \cellcolor{gray!25}{\small \textbf{73.1}} & \cellcolor{gray!25}{\small \textbf{39.7}} & \cellcolor{gray!25}{\small \textbf{47.0}} \\
    \bottomrule[1.5pt]
    \end{tabular}
    \label{tab:main_results}
    \vspace{-0.5em}
    \end{table*}
    
\subsubsection{Incorporating Modality Priors}\label{sec:prior}
While the closed-form fusion weights from Section~\ref{sec:taylor_approx} balance visual and reasoning shifts purely by their magnitudes, MLLM decoders exhibit a clear functional hierarchy: 
Shallow layers (small $l$) predominantly attend to visual tokens, anchoring the model in perceptual representations.
Deep layers (large $l$) focus on textual tokens, supporting abstraction and symbolic reasoning. 

To encode this prior knowledge, we introduce layer-dependent modality priors $w_V^{(l)}, w_R^{(l)}>0$ and reformulate the per-layer fusion objective:
\begin{equation}
\min_{\lambda_V,\lambda_R}\;
    w_V^{(l)}\,\mathrm{LTLD}_V^{(l)}(\lambda_V,\lambda_R)
    + w_R^{(l)}\,\mathrm{LTLD}_R^{(l)}(\lambda_V,\lambda_R).
\end{equation}
  The modality prior-guided closed-form is 
  \begin{equation}\label{eq:modality-prior-closed-form}
    \lambda_t^{(l)}
    = \frac{w_t^{(l)}\,\|\tau_b^{(l)}\|^2}
           {w_V^{(l)}\,\|\tau_V^{(l)}\|^2 + w_R^{(l)}\,\|\tau_R^{(l)}\|^2}, 
    \quad t\in\{V,R\}.
  \end{equation}
    
  \paragraph{Attention-Guided Decay Priors}
During our experiments, we observed that compared to deeper decoder layers, shallower decoder layers in the MLLM allocate more attention to visual tokens, thereby facilitating visual perceptual grounding, while deeper layers focus primarily on textual semantics, as illustrated in Figure~\ref{fig:prior}. 
To introduce reasoning and reflection capabilities while preserving visual perception, and to maintain methodological simplicity without introducing additional supervision, we derive modality priors from the model's layer-wise attention patterns and fit them with an exponential decay function (prioritizing simplicity, though more complex quadratic functions could alternatively be used for fitting). 
This approach naturally captures the non-uniform transition from visual grounding to symbolic reasoning observed in practice. 
Specifically, we first collect the visual attention weights $a_l$ from each decoder layer in the MLLM.
  Then, we posit
    $
   a_l \approx C\,e^{-\hat\alpha \, l},
    $
   and obtain $\hat\alpha,\log C$ via linear regression on
    $\bigl\{(l,\log a_l)\bigr\}_{l=1}^L$. 
  Finally, we set
    \begin{equation}
   w_V^{(l)}
   = \frac{\exp(-\hat\alpha\, l)}{\sum_{j=1}^L \exp(-\hat\alpha\,j)},
   \qquad
   w_R^{(l)} = 1 - w_V^{(l)}.
    \end{equation}
  This attention-guided exponential schedule requires no labels and ensures the modality priors faithfully mirror the model's intrinsic shift from visual grounding to reasoning across the decoder hierarchy. 
  Please refer to Appendix~\ref{app:attention_priors} for details.

  To conclude, FRANK delivers a training-free, interpretable, and efficient framework for the per-layer fusion of visual grounding and logical reasoning in MLLMs, requiring only task-vector norms and simple priors. 
  Its closed-form weights avoid any extra labeling data or optimization, making FRANK practical for scaling multimodal intelligence.

  \begin{table*}[t]
    \centering
    \small
    \renewcommand{\arraystretch}{1.1}
    \captionof{table}{Perceptual performance of FRANK-15B on MME compared to vision-only NVIL-15B. 
    ``w/o MP'' denotes the ablation without Modality Prior. 
    Sub-task abbreviations: ``Comm.'' = Commonsense Reasoning, ``Num.'' = Numerical Calculation, ``Text.'' = Text Translation, ``Code.'' = Code Reasoning.}
    \setlength{\tabcolsep}{5.5pt}
    \begin{tabular}{lccccccc}
        \toprule
        MME               & Existence & Count & Position &  Color & OCR & Poseter & Celebrity  \\
        \midrule
        \cellcolor{gray!25}NVIL-15B (\textbf{\textcolor{red}{upper bound}})    &\cellcolor{gray!25} 100.0  &\cellcolor{gray!25}80.0   &\cellcolor{gray!25}86.7  &\cellcolor{gray!25}91.7   &\cellcolor{gray!25}90.0  &\cellcolor{gray!25}94.2  &\cellcolor{gray!25}83.2   \\
        FRANK-15B w/o MP           & 96.7 & 78.3 & \textbf{68.3} & 90.0 & 75.0 & 91.2 & 76.2  \\
        FRANK-15B            &\textbf{96.7} &\textbf{80.0}  &60.0 &\textbf{91.7}  &\textbf{85.0} &\textbf{91.8} &\textbf{77.7}   \\
        \midrule
        MME               & Scene & Landmark  &  Artwork & Comm. & Num. & Text.  & Code.  \\
        \midrule
        \cellcolor{gray!25}NVIL-15B (\textbf{\textcolor{red}{upper bound}})             &\cellcolor{gray!25}83.5  &\cellcolor{gray!25}90.5   &\cellcolor{gray!25}82.3  &\cellcolor{gray!25}82.9   &\cellcolor{gray!25}72.5  &\cellcolor{gray!25}67.5  &\cellcolor{gray!25}85.0   \\
        FRANK-15B w/o MP           & 80.0 & 83.7 & \textbf{76.5} & 84.3 & \textbf{75.0} & 57.5 & 82.5  \\
        FRANK-15B            &\textbf{83.0} &\textbf{84.3}  &74.3 &\textbf{85.0}  &72.5 &\textbf{60.0} &\textbf{85.0}   \\
        \bottomrule
    \end{tabular}
    \label{tab:MME}
  \end{table*}
  
  \section{Experiments}
  \subsection{Datasets and Model Variants}
  
  We evaluate FRANK on five widely-used multimodal reasoning benchmarks: MMMU val~\cite{YueNZ0LZSJRSWYY24}, MMMU-Pro standard (10 opts)~\cite{yue2024mmmu}, MathVista testmini~\cite{LuBX0LH0CG024}, MathVision testmini~\cite{WangPSLRZZL24}, and WeMath testmini~\cite{qiao2024we}. 
  These datasets encompass a broad spectrum of task formats (e.g., diagram interpretation, symbolic math problems) and difficulty levels.
  
  To study the effects of model scale and architecture, we instantiate three FRANK variants by applying our layer-wise fusion to pairs of non-reasoning MLLMs and reasoning-specialized LLMs. 
  Specifically, \textbf{FRANK-8B} merges Idefics3-8B~\cite{laurenccon2024building} (non-reasoning MLLM) with DeepSeekDistil-LLaMA3-8B~\cite{guo2025deepseek} (reasoning LLM), \textbf{FRANK-15B} merges NVIL-15B~\cite{liu2024nvila} with DeepSeekDistil-Qwen2.5-14B~\cite{guo2025deepseek}, and \textbf{FRANK-38B} merges InternVL2.5-38B~\cite{chen2024expanding} with QwQ-32B~\cite{QwQ}. 
  These variants test our fusion across LLaMA- and Qwen-based architectures and varying capacities. 
  These three FRANK variants enable analysis of our approach across small, medium, and large model scales. 
  We apply FRANK's closed-form fusion weights layer-by-layer without any additional fine-tuning. 
  The attention-guided exponential decay prior is fit on 1000 randomly sampled validation examples from MSCOCO dataset~\cite{LinMBHPRDZ14} (\textbf{note: only the image is needed, no annotations}) using least-squares regression to obtain decay parameter $\alpha$ and normalization constant $C$.
  
  \subsection{Quantitative Evaluation}
  
  Table~\ref{tab:main_results} presents the performance of FRANK variants and state-of-the-art baselines on five multimodal reasoning benchmarks. Below we analyze results by dataset, highlighting both absolute gains and relative improvements to demonstrate the efficacy and scaling behavior of our layer-wise fusion.
  
  \textbf{MMMU/MMMU-Pro.} 
  On MMMU (college-level image-text questions), FRANK-8B achieves 48.3, a 4.4-point gain over its vision branch. 
  This improvement indicates that even at 8B parameters, our fusion effectively combines visual grounding and reasoning. 
  On the more stringent MMMU-Pro, FRANK-8B reaches 34.7. 
  Increasing model capacity yields further gains: FRANK-15B attains 61.3 on MMMU and 49.4 on MMMU-Pro, demonstrating that additional parameters enable richer reasoning adaptations. 
  FRANK-38B further improves to 69.2 (MMMU) and 56.8 (MMMU-Pro), outperforming InternVL2.5-38B by 5.3 and 8.8 points, respectively, and confirming a strong scaling trend.
  \begin{table}[t]
    \begin{minipage}{0.5\textwidth}
    \centering
    \small
    \renewcommand{\arraystretch}{0.8}
    \caption{Ablation study results of different model mergeing methods on the MMMU.} 
    \begin{tabular}{lc}
        \toprule
        \textbf{Method}                         & \textbf{MMMU Acc.} \\
        \midrule
        \cellcolor{gray!25}NVIL-15B (baseline)                              &\cellcolor{gray!25} 53.2             \\
        \midrule
        VLM-Merging                             & 53.6            \\
        Task Arithmetic                        & 56.1            \\
        MetaGPT                                & 57.9            \\
        \midrule
        FRANK-15B w/o Modality Prior & 58.4            \\
        FRANK-15B                       & \textbf{61.3}   \\
        \bottomrule
    \end{tabular}
    \label{tab:ablation}
    \end{minipage}
    \hfill
    \begin{minipage}{0.5\textwidth}
    \centering
    \small
    \caption{The number of reflection tokens in MMMU responses across models.}
    \begin{tabular}{lcc}
      \toprule
      \textbf{Reflection} & \textbf{NVIL-15B} & \textbf{FRANK-15B} \\
      \midrule
      Wait                      & 0               & 20125             \\
      Hmm                       & 0               & 1733            \\
      Mistake                   & 0               & 1663             \\
      Alternatively             & 0               & 7712              \\
      Check                     & 0               & 1586             \\
      \bottomrule
    \end{tabular}
    \label{tab:reflection_transposed}
    \end{minipage}
    \end{table}
  \textbf{Math Benchmarks (MathVista, MathVision, WeMath).}
  We evaluate FRANK variants on three math-focused datasets (Table \ref{tab:main_results}). 
  FRANK-8B and FRANK-15B underperform on MathVista (50.7 and 55.4, -7.7 and -12.2 vs.\ Idefics3-LLaMA3 and NVIL-15B). 
  In contrast, FRANK-38B achieves 73.1 on MathVista (+1.2 vs.\ InternVL2.5), 39.7 on MathVision (+7.5), and 47.0 on WeMath (+8.7), demonstrating that larger capacity better absorbs fusion weights, mitigating model merging interference, while still enhancing deeper-layer symbolic reasoning.
  
  \textbf{Scale and Architecture Analysis.}  
  Aggregating results across all three benchmarks reveals a clear scaling law: FRANK variants continue to reap sustained reasoning benefits and exhibit greater robustness as model size increases. 
  Both Qwen-based FRANK-15B and LLaMA-based FRANK-8B showed consistent improvements in reasoning capabilities, indicating that our layer-wise fusion is independent of the backbone architecture. 
  Moreover, FRANK-38B's strong performance on the most challenging benchmarks underscores that larger models can more fully exploit the injected reasoning capabilities, all achieved without any additional gradient-based training.
  
  \subsection{Visual Perception Evaluation}
  We assess visual perception capability on the MME benchmark~\cite{liang2024survey} (Table~\ref{tab:MME}). 
  NVIL-15B, being a non-reasoning multimodal model, achieves the strong perceptual performance across all 14 sub-tasks, representing the upper bound on visual grounding accuracy. 
  As expected, fusing in decoder layers incurs some degradation; however, by incorporating our attention-guided Decay Prior, FRANK-15B only experiences very slight drops relative to NVIL-15B, substantially outperforming the no-prior ablation. 
  By contrast, the ablated FRANK-15B w/o MP falls further behind, indicating that the Modality Prior is crucial for preserving visual perception when fusing the model. 
  Moreover, in sub-tasks requiring commonsense reasoning (``Comm.''), FRANK-15B even slightly surpasses NVIL-15B (85.0 vs 82.9), suggesting that our fusion maintains and can enhance performance in scenarios where light reasoning complements visual understanding.

  \subsection{Ablation Study}
  Table~\ref{tab:ablation} evaluates each fusion component on MMMU using the same NVIL-15B and DeepSeekDistil-Qwen2.5-14B backbones without fine-tuning. 
  The results confirm that model merging injects reasoning into a pretrained MLLM: 
  \textbf{Traditional fusion baselines} (VLM-Merging~\cite{chen2025bring}, Task Arithmetic~\cite{IlharcoRWSHF23}, and MetaGPT~\cite{ZhouSWC24}) achieve 53.6\%, 56.1\%, and 57.9\%, respectively. 
  Task Arithmetic follows a fixed fusion weight $\lambda_{V}$ of 0.3, shown in prior work~\cite{IlharcoRWSHF23,ZhouSWC24} to perform robustly across diverse tasks, which nonetheless yields only modest gains (3.6-4.7 points), indicating limited integration of reasoning capabilities. 
  \textbf{Layer-only fusion} (FRANK-15B w/o Modality Prior) leverages closed-form weights for per-layer merging and improves to 58.4\%, outperforming all traditional methods by 1.7-3.6 points. 
  \textbf{Full FRANK-15B} adds the attention-guided exponential decay modality prior and attains 61.3\%, a further 2.9-point gain. 
  This confirms that the learned prior successfully balances visual grounding and symbolic reasoning across layers.   

  \begin{figure}[tbp]
    \centering
      \includegraphics[width=0.95\linewidth]{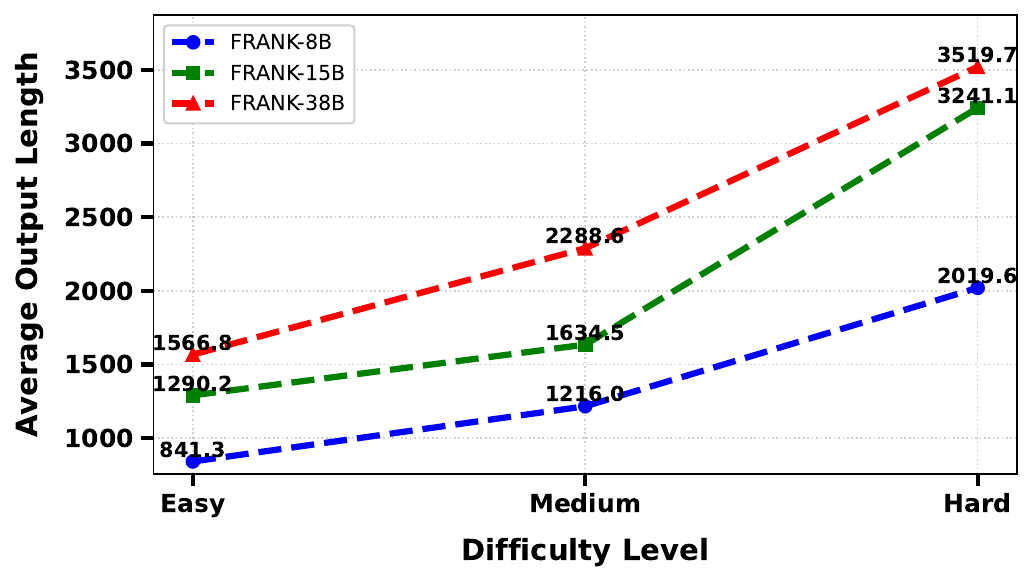}
    \caption{Average output length of the FRANK on the MMMU benchmark, stratified by task difficulty.}
    \label{fig:vis_length}
    \end{figure}

  \subsection{Reflection Frequency and Output Length Analysis}
    We quantify FRANK's self-correction capability by analyzing reflection tokens (\texttt{Wait}, \texttt{Hmm}, \texttt{Mistake}, \texttt{Alternatively}, \texttt{Check}) in MMMU val set responses (N=900). 
    As Table~\ref{tab:reflection_transposed} shows: 
    1) NVIL-15B (non-reasoning) produces zero reflection tokens. 
    2) FRANK-15B generates multiple reflection cycles per example
    This demonstrates that our fusion method intrinsically enables iterative self-correction during reasoning. 
    Figure~\ref{fig:vis_length} reveals two scaling trends on MMMU: 
    1) Output length increases with task difficulty (Easy→Hard). 
    2) Larger models produce longer responses at each level. 
    This demonstrates our method's \textbf{inference scaling} capability, which automatically expands reasoning for harder problems while benefiting from increased model capacity.

\subsection{Case Study: Exemplary Model Outputs}

To illustrate the qualitative improvements brought by our FRANK fusion, we present three representative examples—one for each model size (8B, 15B, 38B)—comparing a non-reasoning baseline against the corresponding FRANK variant. In each case, the baseline model provides a direct (and incorrect) answer, whereas the FRANK model produces a detailed chain-of-thought with reflection tokens, arriving at the correct solution.

\begin{figure*}[htbp]
    \centering
      \includegraphics[width=0.95\textwidth]{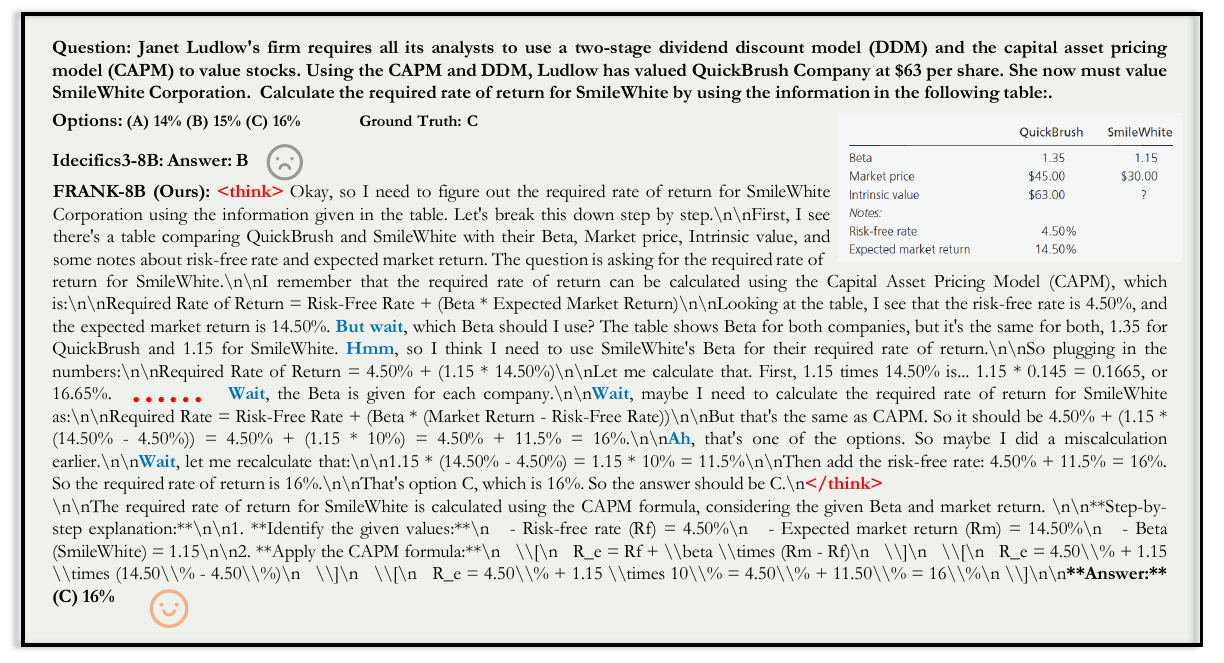}
    \caption{Output examples from FRANK-8B and the non-reasoning baseline model Idecifics3-8B. Here, \textcolor{red}{<think>} and \textcolor{red}{</think>} denote R1-like reasoning processes, while blue text indicates reflection tokens.}
    \label{fig:case-8b}
  \end{figure*}

\begin{figure*}[htbp]
    \centering
      \includegraphics[width=0.95\textwidth]{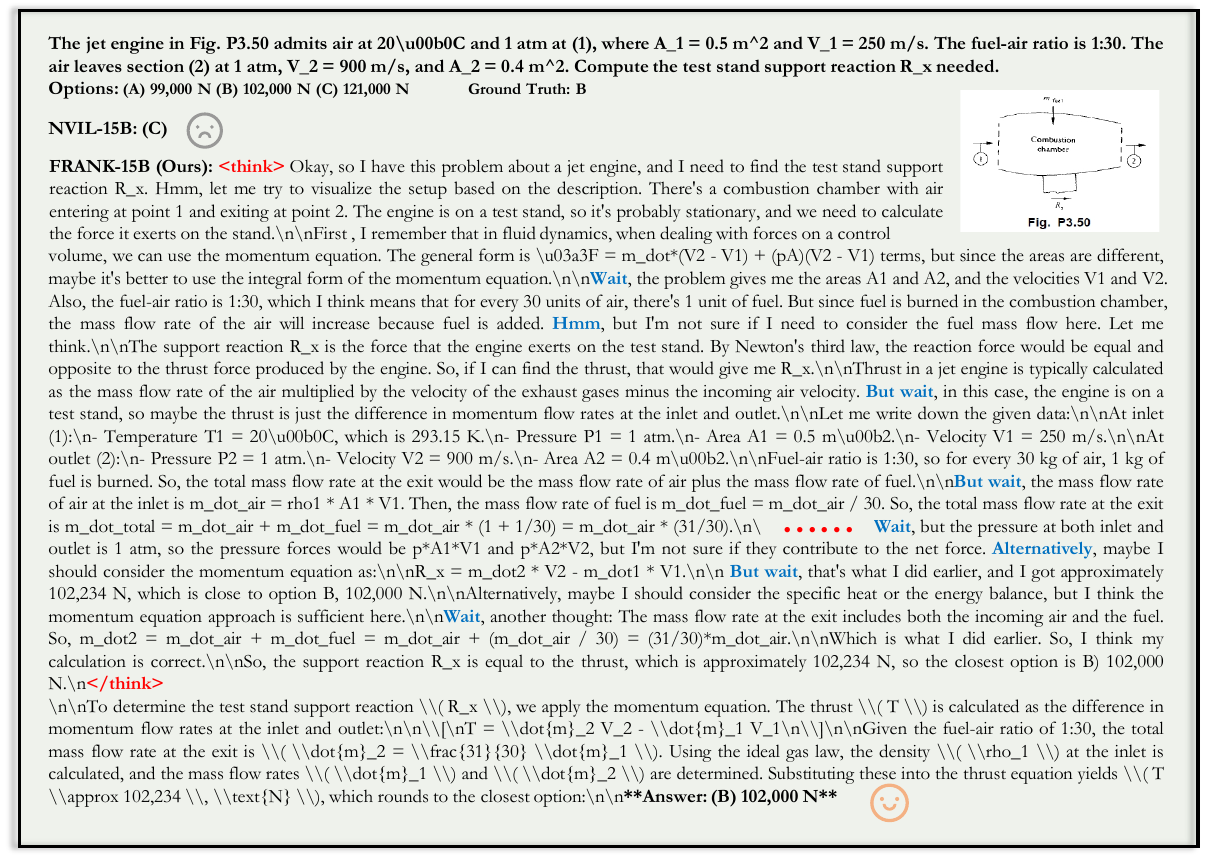}
    \caption{Output examples from FRANK-15B and the non-reasoning baseline model NVIL-15B. }
    \label{fig:case-15b}
  \end{figure*}

\begin{figure*}[htbp]
    \centering
      \includegraphics[width=0.98\textwidth]{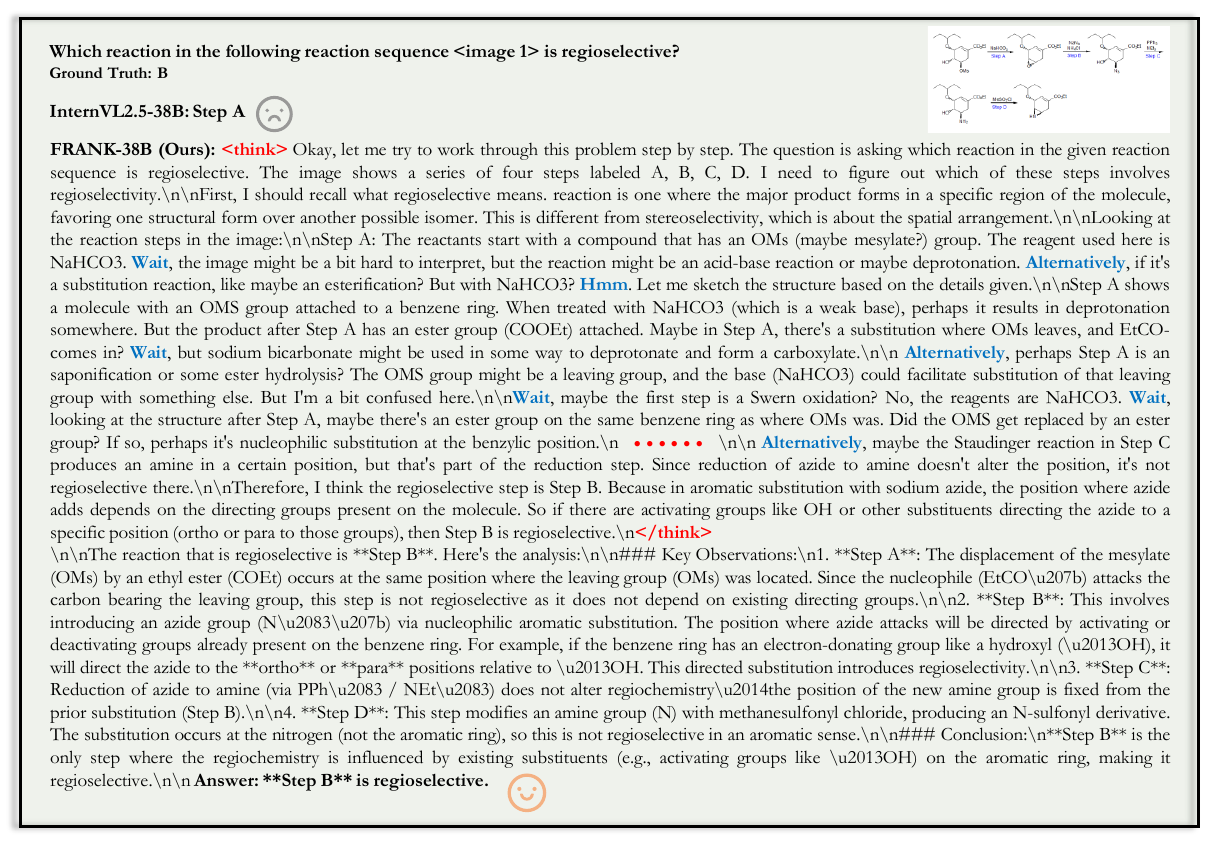}
    \caption{Output examples from FRANK-38B and the non-reasoning baseline model InternVL2.5-38B. }
    \label{fig:case-38b}
  \end{figure*}

As shown in Figures~\ref{fig:case-8b}, ~\ref{fig:case-15b}, and ~\ref{fig:case-38b}, across all three model scales, the FRANK variants consistently produce step-by-step reasoning chains and explicit reflection checks. This enables them to correct intermediate arithmetic or logical reasoning and arrive at the correct results. In contrast, the non-reasoning baselines directly output an incorrect answer without any justification. 

  \section{Conclusions}
  In this paper, we presented FRANK, a training-free and R1-like MLLM that endows off-the-shelf MLLMs with advanced reasoning and self-reflection capabilities. 
  By decomposing the decoder into per-layer task vectors and deriving closed-form fusion weights under NTK linearization and task-vector orthogonality, FRANK seamlessly integrates visual grounding from vision-fine-tuned MLLMs with logical reasoning from reasoning-specialized LLMs, guided by attention-driven modality priors. 
  Extensive experiments across five multimodal reasoning benchmarks demonstrate that FRANK consistently outperforms state-of-the-art baselines at 8B, 15B, and 38B scales, with particularly strong gains on reasoning tasks. 

  Future work includes extending FRANK to support more diverse modalities (e.g., audio, video), exploring dynamic fusion strategies for real-time tasks, and investigating theoretical guarantees under broader neural architectures. We believe FRANK offers a practical and interpretable path toward scalable multimodal intelligence without the overhead of task-specific retraining.

\bibliography{ms}

\begin{thebibliography}{53}
\providecommand{\natexlab}[1]{#1}
\providecommand{\url}[1]{\texttt{#1}}
\expandafter\ifx\csname urlstyle\endcsname\relax
  \providecommand{\doi}[1]{doi: #1}\else
  \providecommand{\doi}{doi: \begingroup \urlstyle{rm}\Url}\fi

\bibitem[Guo et~al.(2025)Guo, Yang, Zhang, Song, Zhang, Xu, Zhu, Ma, Wang, Bi, et~al.]{guo2025deepseek}
Daya Guo, Dejian Yang, Haowei Zhang, Junxiao Song, Ruoyu Zhang, Runxin Xu, Qihao Zhu, Shirong Ma, Peiyi Wang, Xiao Bi, et~al.
\newblock {DeepSeek-R1}: Incentivizing reasoning capability in llms via reinforcement learning.
\newblock \emph{arXiv preprint arXiv:2501.12948}, 2025.

\bibitem[Team et~al.(2025)Team, Du, Gao, Xing, Jiang, Chen, Li, Xiao, Du, Liao, et~al.]{team2025kimi}
Kimi Team, Angang Du, Bofei Gao, Bowei Xing, Changjiu Jiang, Cheng Chen, Cheng Li, Chenjun Xiao, Chenzhuang Du, Chonghua Liao, et~al.
\newblock {Kimi K1.5}: Scaling reinforcement learning with llms.
\newblock \emph{arXiv preprint arXiv:2501.12599}, 2025.

\bibitem[Ma et~al.(2025)Ma, Wang, Liu, Liu, Chen, Zhang, Zhou, Du, and Li]{ma2025s}
Ruotian Ma, Peisong Wang, Cheng Liu, Xingyan Liu, Jiaqi Chen, Bang Zhang, Xin Zhou, Nan Du, and Jia Li.
\newblock {S $^2$ R}: Teaching llms to self-verify and self-correct via reinforcement learning.
\newblock \emph{arXiv preprint arXiv:2502.12853}, 2025.

\bibitem[Lightman et~al.(2024)Lightman, Kosaraju, Burda, Edwards, Baker, Lee, Leike, Schulman, Sutskever, and Cobbe]{LightmanKBEBLLS24}
Hunter Lightman, Vineet Kosaraju, Yuri Burda, Harrison Edwards, Bowen Baker, Teddy Lee, Jan Leike, John Schulman, Ilya Sutskever, and Karl Cobbe.
\newblock Let's verify step by step.
\newblock In \emph{ICLR}, 2024.

\bibitem[{OpenAI}(2024{\natexlab{a}})]{OpenAI}
{OpenAI}.
\newblock {OpenAI} o1.
\newblock https://openai.com/o1/, 2024{\natexlab{a}}.

\bibitem[Xu et~al.(2024)Xu, Jin, Hao, Song, Sun, and Yuan]{xu2024llava}
Guowei Xu, Peng Jin, Li~Hao, Yibing Song, Lichao Sun, and Li~Yuan.
\newblock {LLaVA-o1}: Let vision language models reason step-by-step.
\newblock \emph{arXiv preprint arXiv:2411.10440}, 2024.

\bibitem[Dong et~al.(2025)Dong, Liu, Sun, Yang, Hu, Rao, and Liu]{dong2024insight}
Yuhao Dong, Zuyan Liu, Hai-Long Sun, Jingkang Yang, Winston Hu, Yongming Rao, and Ziwei Liu.
\newblock {Insight-V}: Exploring long-chain visual reasoning with multimodal large language models.
\newblock In \emph{CVPR}, 2025.

\bibitem[Yao et~al.(2024)Yao, Huang, Wu, Zhang, Wang, Liu, Wang, Song, Feng, Shen, et~al.]{yao2024mulberry}
Huanjin Yao, Jiaxing Huang, Wenhao Wu, Jingyi Zhang, Yibo Wang, Shunyu Liu, Yingjie Wang, Yuxin Song, Haocheng Feng, Li~Shen, et~al.
\newblock Mulberry: Empowering mllm with o1-like reasoning and reflection via collective monte carlo tree search.
\newblock \emph{arXiv preprint arXiv:2412.18319}, 2024.

\bibitem[Liu et~al.(2025)Liu, Sun, Zang, Dong, Cao, Duan, Lin, and Wang]{liu2025visual}
Ziyu Liu, Zeyi Sun, Yuhang Zang, Xiaoyi Dong, Yuhang Cao, Haodong Duan, Dahua Lin, and Jiaqi Wang.
\newblock {Visual-RFT}: Visual reinforcement fine-tuning.
\newblock \emph{arXiv preprint arXiv:2503.01785}, 2025.

\bibitem[Huang et~al.(2025)Huang, Jia, Zhai, Cao, Ye, Zhao, Hu, and Lin]{huang2025vision}
Wenxuan Huang, Bohan Jia, Zijie Zhai, Shaosheng Cao, Zheyu Ye, Fei Zhao, Yao Hu, and Shaohui Lin.
\newblock {Vision-R1}: Incentivizing reasoning capability in multimodal large language models.
\newblock \emph{arXiv preprint arXiv:2503.06749}, 2025.

\bibitem[Peng et~al.(2025)Peng, Zhang, Zhang, You, Liu, Zhu, Yang, Xu, Geng, and Yang]{peng2025lmm}
Yingzhe Peng, Gongrui Zhang, Miaosen Zhang, Zhiyuan You, Jie Liu, Qipeng Zhu, Kai Yang, Xingzhong Xu, Xin Geng, and Xu~Yang.
\newblock {LMM-R1}: Empowering 3b lmms with strong reasoning abilities through two-stage rule-based rl.
\newblock \emph{arXiv preprint arXiv:2503.07536}, 2025.

\bibitem[Zhang et~al.(2025)Zhang, Huang, Yao, Liu, Zhang, Lu, and Tao]{zhang2025r1}
Jingyi Zhang, Jiaxing Huang, Huanjin Yao, Shunyu Liu, Xikun Zhang, Shijian Lu, and Dacheng Tao.
\newblock {R1-VL}: Learning to reason with multimodal large language models via step-wise group relative policy optimization.
\newblock \emph{arXiv preprint arXiv:2503.12937}, 2025.

\bibitem[Wortsman et~al.(2022)Wortsman, Ilharco, Gadre, Roelofs, Lopes, Morcos, Namkoong, Farhadi, Carmon, Kornblith, and Schmidt]{WortsmanIGRLMNF22}
Mitchell Wortsman, Gabriel Ilharco, Samir~Yitzhak Gadre, Rebecca Roelofs, Raphael~Gontijo Lopes, Ari~S. Morcos, Hongseok Namkoong, Ali Farhadi, Yair Carmon, Simon Kornblith, and Ludwig Schmidt.
\newblock Model soups: averaging weights of multiple fine-tuned models improves accuracy without increasing inference time.
\newblock In \emph{ICML}, volume 162, pages 23965--23998, 2022.

\bibitem[Ilharco et~al.(2023)Ilharco, Ribeiro, Wortsman, Schmidt, Hajishirzi, and Farhadi]{IlharcoRWSHF23}
Gabriel Ilharco, Marco~T{\'{u}}lio Ribeiro, Mitchell Wortsman, Ludwig Schmidt, Hannaneh Hajishirzi, and Ali Farhadi.
\newblock Editing models with task arithmetic.
\newblock In \emph{ICLR}, 2023.

\bibitem[Zhang et~al.(2023)Zhang, Chen, Liu, and He]{ZhangCLH23}
Jinghan Zhang, Shiqi Chen, Junteng Liu, and Junxian He.
\newblock Composing parameter-efficient modules with arithmetic operation.
\newblock In \emph{NeurIPS}, New Orleans, LA, USA, 2023.

\bibitem[Brincat et~al.(2018)Brincat, Siegel, von Nicolai, and Miller]{brincat2018gradual}
Scott~L Brincat, Markus Siegel, Constantin von Nicolai, and Earl~K Miller.
\newblock Gradual progression from sensory to task-related processing in cerebral cortex.
\newblock \emph{Proceedings of the National Academy of Sciences}, 115\penalty0 (30):\penalty0 E7202--E7211, 2018.

\bibitem[Kawasaki et~al.(2022)Kawasaki, Nishida, and Kobayashi]{kawasaki2022hierarchical}
Haruka Kawasaki, Satoshi Nishida, and Ichiro Kobayashi.
\newblock Hierarchical processing of visual and language information in the brain.
\newblock In \emph{AACL-IJCNLP}, pages 405--410, 2022.

\bibitem[Wang et~al.(2023)Wang, Chen, Chen, Wu, Zhu, Zeng, Luo, Lu, Zhou, Qiao, and Dai]{VisionLLM}
Wenhai Wang, Zhe Chen, Xiaokang Chen, Jiannan Wu, Xizhou Zhu, Gang Zeng, Ping Luo, Tong Lu, Jie Zhou, Yu~Qiao, and Jifeng Dai.
\newblock Visionllm: Large language model is also an open-ended decoder for vision-centric tasks.
\newblock In \emph{NeurIPS}, 2023.

\bibitem[Bai et~al.(2023)Bai, Bai, Chu, Cui, Dang, Deng, Fan, Ge, Han, Huang, et~al.]{bai2023qwen}
Jinze Bai, Shuai Bai, Yunfei Chu, Zeyu Cui, Kai Dang, Xiaodong Deng, Yang Fan, Wenbin Ge, Yu~Han, Fei Huang, et~al.
\newblock Qwen technical report.
\newblock \emph{arXiv preprint arXiv:2309.16609}, 2023.

\bibitem[Liu et~al.(2023)Liu, Li, Wu, and Lee]{LiuLWL23a}
Haotian Liu, Chunyuan Li, Qingyang Wu, and Yong~Jae Lee.
\newblock Visual instruction tuning.
\newblock In \emph{NeurIPS}, New Orleans, LA, USA, 2023.

\bibitem[Sun et~al.(2024)Sun, Yu, Cui, Zhang, Zhang, Wang, Gao, Liu, Huang, and Wang]{SunYCZZWGL0W24}
Quan Sun, Qiying Yu, Yufeng Cui, Fan Zhang, Xiaosong Zhang, Yueze Wang, Hongcheng Gao, Jingjing Liu, Tiejun Huang, and Xinlong Wang.
\newblock Emu: Generative pretraining in multimodality.
\newblock In \emph{ICLR}, Vienna, Austria, 2024.

\bibitem[Wei et~al.(2022)Wei, Wang, Schuurmans, Bosma, Ichter, Xia, Chi, Le, and Zhou]{Wei0SBIXCLZ22}
Jason Wei, Xuezhi Wang, Dale Schuurmans, Maarten Bosma, Brian Ichter, Fei Xia, Ed~H. Chi, Quoc~V. Le, and Denny Zhou.
\newblock Chain-of-thought prompting elicits reasoning in large language models.
\newblock In \emph{NeurIPS}, New Orleans, LA, USA, 2022.

\bibitem[Zhang et~al.(2024)Zhang, Zhang, Li, Zhang, Sun, Gan, Yang, Pang, and Yang]{zhang2024improve}
Ruohong Zhang, Bowen Zhang, Yanghao Li, Haotian Zhang, Zhiqing Sun, Zhe Gan, Yinfei Yang, Ruoming Pang, and Yiming Yang.
\newblock Improve vision language model chain-of-thought reasoning.
\newblock \emph{arXiv preprint arXiv:2410.16198}, 2024.

\bibitem[Mitra et~al.(2024)Mitra, Huang, Darrell, and Herzig]{MitraHDH24}
Chancharik Mitra, Brandon Huang, Trevor Darrell, and Roei Herzig.
\newblock Compositional chain-of-thought prompting for large multimodal models.
\newblock In \emph{CVPR}, pages 14420--14431, Seattle, WA, USA, 2024.

\bibitem[Luan et~al.(2024)Luan, Feng, Chen, Wang, Zhou, and Li]{luan2024textcot}
Bozhi Luan, Hao Feng, Hong Chen, Yonghui Wang, Wengang Zhou, and Houqiang Li.
\newblock {TextCoT}: Zoom in for enhanced multimodal text-rich image understanding.
\newblock \emph{arXiv preprint arXiv:2404.09797}, 2024.

\bibitem[Meng et~al.(2025)Meng, Du, Liu, Zhou, Lu, Fu, Han, Shi, Wang, He, et~al.]{meng2025mm}
Fanqing Meng, Lingxiao Du, Zongkai Liu, Zhixiang Zhou, Quanfeng Lu, Daocheng Fu, Tiancheng Han, Botian Shi, Wenhai Wang, Junjun He, et~al.
\newblock {MM-Eureka}: Exploring the frontiers of multimodal reasoning with rule-based reinforcement learning.
\newblock \emph{arXiv preprint arXiv:2503.07365}, 2025.

\bibitem[Matena and Raffel(2022)]{MatenaR22}
Michael Matena and Colin Raffel.
\newblock Merging models with fisher-weighted averaging.
\newblock In \emph{NeurIPS}, New Orleans, LA, USA, 2022.

\bibitem[Yadav et~al.(2023)Yadav, Tam, Choshen, Raffel, and Bansal]{YadavTCRB23}
Prateek Yadav, Derek Tam, Leshem Choshen, Colin~A. Raffel, and Mohit Bansal.
\newblock Ties-merging: Resolving interference when merging models.
\newblock In \emph{NeurIPS}, New Orleans, LA, USA, 2023.

\bibitem[Zhou et~al.(2024)Zhou, Song, Wang, and Chen]{ZhouSWC24}
Yuyan Zhou, Liang Song, Bingning Wang, and Weipeng Chen.
\newblock Metagpt: Merging large language models using model exclusive task arithmetic.
\newblock In \emph{EMNLP}, pages 1711--1724, Miami, FL, USA, 2024.

\bibitem[Jin et~al.(2023)Jin, Ren, Preotiuc{-}Pietro, and Cheng]{Jin0P023}
Xisen Jin, Xiang Ren, Daniel Preotiuc{-}Pietro, and Pengxiang Cheng.
\newblock Dataless knowledge fusion by merging weights of language models.
\newblock In \emph{ICLR}, Kigali, Rwanda, 2023.

\bibitem[Yu et~al.(2024)Yu, Yu, Yu, Huang, and Li]{Yu0Y0L24}
Le~Yu, Bowen Yu, Haiyang Yu, Fei Huang, and Yongbin Li.
\newblock Language models are super mario: Absorbing abilities from homologous models as a free lunch.
\newblock In \emph{ICML}, Vienna, Austria, 2024.

\bibitem[Chen et~al.(2025)Chen, Zhang, Zhu, Liu, Gao, Xiong, Li, and He]{chen2025bring}
Shiqi Chen, Jinghan Zhang, Tongyao Zhu, Wei Liu, Siyang Gao, Miao Xiong, Manling Li, and Junxian He.
\newblock Bring reason to vision: Understanding perception and reasoning through model merging.
\newblock \emph{arXiv preprint arXiv:2505.05464}, 2025.

\bibitem[Jacot et~al.(2018)Jacot, Hongler, and Gabriel]{JacotHG18}
Arthur Jacot, Cl{\'{e}}ment Hongler, and Franck Gabriel.
\newblock {Neural Tangent Kernel}: Convergence and generalization in neural networks.
\newblock In \emph{NeurIPS}, pages 8580--8589, Montreal, Canada, 2018.

\bibitem[Touvron et~al.(2023)Touvron, Martin, Stone, Albert, Almahairi, Babaei, Bashlykov, Batra, Bhargava, Bhosale, et~al.]{touvron2023llama-2}
Hugo Touvron, Louis Martin, Kevin Stone, Peter Albert, Amjad Almahairi, Yasmine Babaei, Nikolay Bashlykov, Soumya Batra, Prajjwal Bhargava, Shruti Bhosale, et~al.
\newblock Llama 2: Open foundation and fine-tuned chat models.
\newblock \emph{arXiv preprint arXiv:2307.09288}, 2023.

\bibitem[Zhong et~al.(2024)Zhong, Cui, Guo, Liang, Lu, Wang, Saied, Chen, and Duan]{ZhongCGLLWSCD24}
Wanjun Zhong, Ruixiang Cui, Yiduo Guo, Yaobo Liang, Shuai Lu, Yanlin Wang, Amin Saied, Weizhu Chen, and Nan Duan.
\newblock {AGIEval}: {A} human-centric benchmark for evaluating foundation models.
\newblock In \emph{NAACL}, pages 2299--2314, Mexico City, Mexico, 2024.

\bibitem[Liu et~al.(2024{\natexlab{a}})Liu, Li, Li, and Lee]{LiuLLL24}
Haotian Liu, Chunyuan Li, Yuheng Li, and Yong~Jae Lee.
\newblock Improved baselines with visual instruction tuning.
\newblock In \emph{CVPR}, pages 26286--26296, Seattle, WA, USA, 2024{\natexlab{a}}.

\bibitem[Liu et~al.(2024{\natexlab{b}})Liu, Li, Li, Li, Zhang, Shen, and Lee]{LLaVA-NeXT}
Haotian Liu, Chunyuan Li, Yuheng Li, Bo~Li, Yuanhan Zhang, Sheng Shen, and Yong~Jae Lee.
\newblock Llava-next.
\newblock https://llava-vl.github.io/blog/2024-01-30-llava-next/, 2024{\natexlab{b}}.

\bibitem[Lin et~al.(2024)Lin, Yin, Ping, Molchanov, Shoeybi, and Han]{LinYP0SH24}
Ji~Lin, Hongxu Yin, Wei Ping, Pavlo Molchanov, Mohammad Shoeybi, and Song Han.
\newblock {VILA:} on pre-training for visual language models.
\newblock In \emph{CVPR}, pages 26679--26689, Seattle, WA, USA, 2024.

\bibitem[Lauren{\c{c}}on et~al.(2024)Lauren{\c{c}}on, Marafioti, Sanh, and Tronchon]{laurenccon2024building}
Hugo Lauren{\c{c}}on, Andr{\'e}s Marafioti, Victor Sanh, and L{\'e}o Tronchon.
\newblock Building and better understanding vision-language models: insights and future directions.
\newblock In \emph{Workshop on RBFM}, 2024.

\bibitem[Chen et~al.(2024{\natexlab{a}})Chen, Li, Dong, Zhang, He, Wang, Zhao, and Lin]{ChenLDZHWZL24}
Lin Chen, Jinsong Li, Xiaoyi Dong, Pan Zhang, Conghui He, Jiaqi Wang, Feng Zhao, and Dahua Lin.
\newblock Sharegpt4v: Improving large multi-modal models with better captions.
\newblock In \emph{ECCV}, volume 15075, pages 370--387, Milan, Italy, 2024{\natexlab{a}}.

\bibitem[Liu et~al.(2024{\natexlab{c}})Liu, Zhu, Shi, Zhang, Lou, Yang, Xi, Cao, Gu, Li, et~al.]{liu2024nvila}
Zhijian Liu, Ligeng Zhu, Baifeng Shi, Zhuoyang Zhang, Yuming Lou, Shang Yang, Haocheng Xi, Shiyi Cao, Yuxian Gu, Dacheng Li, et~al.
\newblock Nvila: Efficient frontier visual language models.
\newblock \emph{arXiv preprint arXiv:2412.04468}, 2024{\natexlab{c}}.

\bibitem[Li et~al.(2024)Li, Zhang, Guo, Zhang, Li, Zhang, Zhang, Zhang, Li, Liu, and Li]{li2024llava}
Bo~Li, Yuanhan Zhang, Dong Guo, Renrui Zhang, Feng Li, Hao Zhang, Kaichen Zhang, Peiyuan Zhang, Yanwei Li, Ziwei Liu, and Chunyuan Li.
\newblock {LLaVA-OneVision}: Easy visual task transfer.
\newblock \emph{TMLR}, 2024.

\bibitem[Wang et~al.(2024{\natexlab{a}})Wang, Bai, Tan, Wang, Fan, Bai, Chen, Liu, Wang, Ge, et~al.]{wang2024qwen2}
Peng Wang, Shuai Bai, Sinan Tan, Shijie Wang, Zhihao Fan, Jinze Bai, Keqin Chen, Xuejing Liu, Jialin Wang, Wenbin Ge, et~al.
\newblock {Qwen2-VL}: Enhancing vision-language model's perception of the world at any resolution.
\newblock \emph{arXiv preprint arXiv:2409.12191}, 2024{\natexlab{a}}.

\bibitem[{OpenAI}(2024{\natexlab{b}})]{GPT-4o}
{OpenAI}.
\newblock {GPT-4o}.
\newblock https://openai.com/index/hello-gpt-4o/, 2024{\natexlab{b}}.

\bibitem[Chen et~al.(2024{\natexlab{b}})Chen, Wang, Cao, Liu, Gao, Cui, Zhu, Ye, Tian, Liu, et~al.]{chen2024expanding}
Zhe Chen, Weiyun Wang, Yue Cao, Yangzhou Liu, Zhangwei Gao, Erfei Cui, Jinguo Zhu, Shenglong Ye, Hao Tian, Zhaoyang Liu, et~al.
\newblock Expanding performance boundaries of open-source multimodal models with model, data, and test-time scaling.
\newblock \emph{arXiv preprint arXiv:2412.05271}, 2024{\natexlab{b}}.

\bibitem[Yue et~al.(2024{\natexlab{a}})Yue, Ni, Zheng, Zhang, Liu, Zhang, Stevens, Jiang, Ren, Sun, Wei, Yu, Yuan, Sun, Yin, Zheng, Yang, Liu, Huang, Sun, Su, and Chen]{YueNZ0LZSJRSWYY24}
Xiang Yue, Yuansheng Ni, Tianyu Zheng, Kai Zhang, Ruoqi Liu, Ge~Zhang, Samuel Stevens, Dongfu Jiang, Weiming Ren, Yuxuan Sun, Cong Wei, Botao Yu, Ruibin Yuan, Renliang Sun, Ming Yin, Boyuan Zheng, Zhenzhu Yang, Yibo Liu, Wenhao Huang, Huan Sun, Yu~Su, and Wenhu Chen.
\newblock {MMMU:} {A} massive multi-discipline multimodal understanding and reasoning benchmark for expert {AGI}.
\newblock In \emph{CVPR}, pages 9556--9567, Seattle, WA, USA, 2024{\natexlab{a}}.

\bibitem[Yue et~al.(2024{\natexlab{b}})Yue, Zheng, Ni, Wang, Zhang, Tong, Sun, Yu, Zhang, Sun, et~al.]{yue2024mmmu}
Xiang Yue, Tianyu Zheng, Yuansheng Ni, Yubo Wang, Kai Zhang, Shengbang Tong, Yuxuan Sun, Botao Yu, Ge~Zhang, Huan Sun, et~al.
\newblock {MMMU-Pro}: {A} more robust multi-discipline multimodal understanding benchmark.
\newblock \emph{arXiv preprint arXiv:2409.02813}, 2024{\natexlab{b}}.

\bibitem[Lu et~al.(2024)Lu, Bansal, Xia, Liu, Li, Hajishirzi, Cheng, Chang, Galley, and Gao]{LuBX0LH0CG024}
Pan Lu, Hritik Bansal, Tony Xia, Jiacheng Liu, Chunyuan Li, Hannaneh Hajishirzi, Hao Cheng, Kai{-}Wei Chang, Michel Galley, and Jianfeng Gao.
\newblock {MathVista}: Evaluating mathematical reasoning of foundation models in visual contexts.
\newblock In \emph{ICLR}, Vienna, Austria, 2024.

\bibitem[Wang et~al.(2024{\natexlab{b}})Wang, Pan, Shi, Lu, Ren, Zhou, Zhan, and Li]{WangPSLRZZL24}
Ke~Wang, Junting Pan, Weikang Shi, Zimu Lu, Houxing Ren, Aojun Zhou, Mingjie Zhan, and Hongsheng Li.
\newblock Measuring multimodal mathematical reasoning with math-vision dataset.
\newblock In \emph{NeurIPS}, Vancouver, BC, Canada, 2024{\natexlab{b}}.

\bibitem[Qiao et~al.(2024)Qiao, Tan, Dong, Wu, Sun, Song, GongQue, Lei, Wei, Zhang, et~al.]{qiao2024we}
Runqi Qiao, Qiuna Tan, Guanting Dong, Minhui Wu, Chong Sun, Xiaoshuai Song, Zhuoma GongQue, Shanglin Lei, Zhe Wei, Miaoxuan Zhang, et~al.
\newblock We-math: Does your large multimodal model achieve human-like mathematical reasoning?
\newblock \emph{arXiv preprint arXiv:2407.01284}, 2024.

\bibitem[Qwen(2024)]{QwQ}
Qwen.
\newblock Qwq.
\newblock https://qwenlm.github.io/blog/qwq-32b-preview/, 2024.

\bibitem[Lin et~al.(2014)Lin, Maire, Belongie, Hays, Perona, Ramanan, Doll{\'{a}}r, and Zitnick]{LinMBHPRDZ14}
Tsung{-}Yi Lin, Michael Maire, Serge~J. Belongie, James Hays, Pietro Perona, Deva Ramanan, Piotr Doll{\'{a}}r, and C.~Lawrence Zitnick.
\newblock Microsoft {COCO:} common objects in context.
\newblock In \emph{ECCV}, volume 8693, pages 740--755, Zurich, Switzerland, 2014.

\bibitem[Liang et~al.(2024)Liang, Xu, Hong, Shang, Wang, Fu, and Liu]{liang2024survey}
Zijing Liang, Yanjie Xu, Yifan Hong, Penghui Shang, Qi~Wang, Qiang Fu, and Ke~Liu.
\newblock A survey of multimodel large language models.
\newblock In \emph{CAICE}, pages 405--409, 2024.

\end{thebibliography}
\clearpage
  \begin{table*}[tbp]
    \centering
    \caption{Detailed fusion components for FRANK configurations}
    \label{tab:implementation_details}
    \footnotesize
    \begin{tabular}{lccc}
    \toprule
    \textbf{FRANK Variant} & \textbf{Non-Reasoning MLLM} & \textbf{Reasoning LLM} & \textbf{Base Model} \\
    \midrule
    FRANK-8B & Idefics3-8B & DeepSeekDistil-LLaMA3-8B & LLaMA3.1-8B \\
    FRANK-15B & NVIL-15B & DeepSeekDistil-Qwen2.5-14B & Qwen2.5-14B \\
    FRANK-38B & InternVL2.5-38B & QwQ-32B & Qwen2.5-32B \\
    \bottomrule
    \end{tabular}
    \end{table*}
\appendix
  \section{Appendix}
  \subsection{Implementation Details}

    As shown in Table~\ref{tab:implementation_details}, we summarize the precise fusion configurations used in our FRANK variant. 
    For each model size, we list the vision-finetuned model, the reasoning-finetuned model, and the underlying base model that we linearly combine via our layer-wise Taylor-derived weights. 
    All our experiments were conducted on NVIDIA GPUs.

    In each case, the ``Non-Reasoning MLLM'' provides the task vector encoding of the vision-adapted decoder updates, the ``Reasoning LLM'' supplies the task vector for pure language reasoning, and the ``Base Model'' is the pre-trained backbone into which these vectors are fused according to our attention-guided exponential decay schedule.

    \subsection{Ablation of the Multimodal Prior}
    \begin{table*}[tbp]
        \captionsetup{font={footnotesize}}
        \caption{Ablation study results of FRANK-15B across five multimodal reasoning benchmarks: MMMU val, MMMU-Pro standard (10 opts), MathVista testmini, MathVision testmini, and WeMath testmini. 
        * indicates the baseline model.}
        \centering
        \footnotesize
        \renewcommand{\arraystretch}{1} 
        \setlength{\tabcolsep}{1.9pt} 
        \begin{tabular}{m{0.5cm}|l|cccccc}
        \toprule[1.6pt]
        & {\footnotesize \textbf{Methods}} & {\footnotesize \textbf{MMMU}} & {\footnotesize \textbf{MMMU-Pro}} & {\footnotesize \textbf{MathVista}} & {\footnotesize {\textbf{MathVision}}} & {\footnotesize \textbf{WeMath}} \\ 
        \midrule[0.7pt]
        & {\small NVILA*} & {\small 53.2} & {\small 36.2} & {\small \textbf{67.6}} & {\small 23.2} & {\small 31.1} \\
        & {\small FRANK-15B w/o MP } & {\small 58.4} & {\small 41.8} & {\small 52.5} & {\small 28.7} & {\small 31.4} \\
        & \cellcolor{gray!25}{\small \textbf{FRANK-15B}} & \cellcolor{gray!25}{\small \textbf{61.3}} & \cellcolor{gray!25}{\small \textbf{49.4}} & \cellcolor{gray!25}{\small {55.4}} & \cellcolor{gray!25}{\small \textbf{37.2}} & \cellcolor{gray!25}{\small \textbf{32.3}} \\
        \bottomrule[1.5pt]
        \end{tabular}
        \label{tab:ablation_results}
        \vspace{-0.5em}
    \end{table*}
    To validate the contribution of our attention-driven multimodal prior (MP), we perform an ablation study on FRANK-15B across five standard benchmarks. 
    Table~\ref{tab:ablation_results} compares the full FRANK-15B against a variant with the MP component removed (w/o MP). 
    On MMMU, accuracy drops from 61.3\% to 58.4\% when MP is disabled, a 2.9-point decrease that underscores the prior's role in balancing visual and textual signals. 
    The impact is even more pronounced on MathVision, where performance falls from 37.2\% to 28.7\%, reflecting a 8.5-point degradation in complex visual-mathematical reasoning. 
    We observe consistent drops on the remaining benchmarks (MMMU-Pro, MathVista, and WeMath), indicating that MP systematically guides layer-wise fusion towards more effective cross-modal integration. 
    These results confirm that our multimodal prior is essential for robust reasoning: by weighting visual and language contributions according to per-layer attention, it ensures that FRANK-15B can harness both modalities optimally across diverse tasks.

  \begin{table}[tbp]
    \centering
    \caption{Ablation study results of  attention-guided exponential decay prior on the MMMU.}
        \label{tab:decay-ablation} 
        \begin{tabular}{lc}
            \toprule
            \textbf{Method}                  & \textbf{MMMU Acc.} \\
            \midrule
            \cellcolor{gray!25}NVIL-15B (baseline) & \cellcolor{gray!25} 53.2 \\
            \midrule
            0.3                     & 59.1 \\
            0.5                 & 56.3 \\
            linear                        & 59.6 \\
            \midrule
            FRANK-15B                         & \textbf{61.3} \\
            \bottomrule
        \end{tabular}
    \end{table}

    \subsection{Ablation of the Attention-Guided Exponential Decay Prior}
    To further assess the effectiveness of our attention-guided exponential decay strategy, we compare against two manually chosen exponential decay factors (0.3 and 0.5) and a simple layer-indexed linear decay. 
    As shown in Table~\ref{tab:decay-ablation}, both fixed-rate exponentials (59.1\% and 56.3\%) and the linear schedule (59.6\%) improve over the NVIL-15B baseline (53.2\%), but remain significantly below our full FRANK-15B (61.3\%). 
    This clear margin demonstrates that leveraging per-layer visual attention to drive the decay adapts the modality weighting more precisely than uniform schedules and achieves the best trade-off between visual perception and language reasoning. 

    \subsection{Validation of Layer-Wise Multimodal Prior and Task-Vector Orthogonality}

    To verify the two key assumptions underpinning our fusion strategy—namely that (1) shallow decoder layers focus predominantly on visual perception while deeper layers prioritize language reasoning (the multimodal prior), and (2) task vectors from vision-finetuned and reasoning-finetuned models are mutually orthogonal—we analyze both Idefics3-8B and InternVL2.5-38B.

\begin{figure}[tbp]
    \centering
      \includegraphics[width=0.45\textwidth]{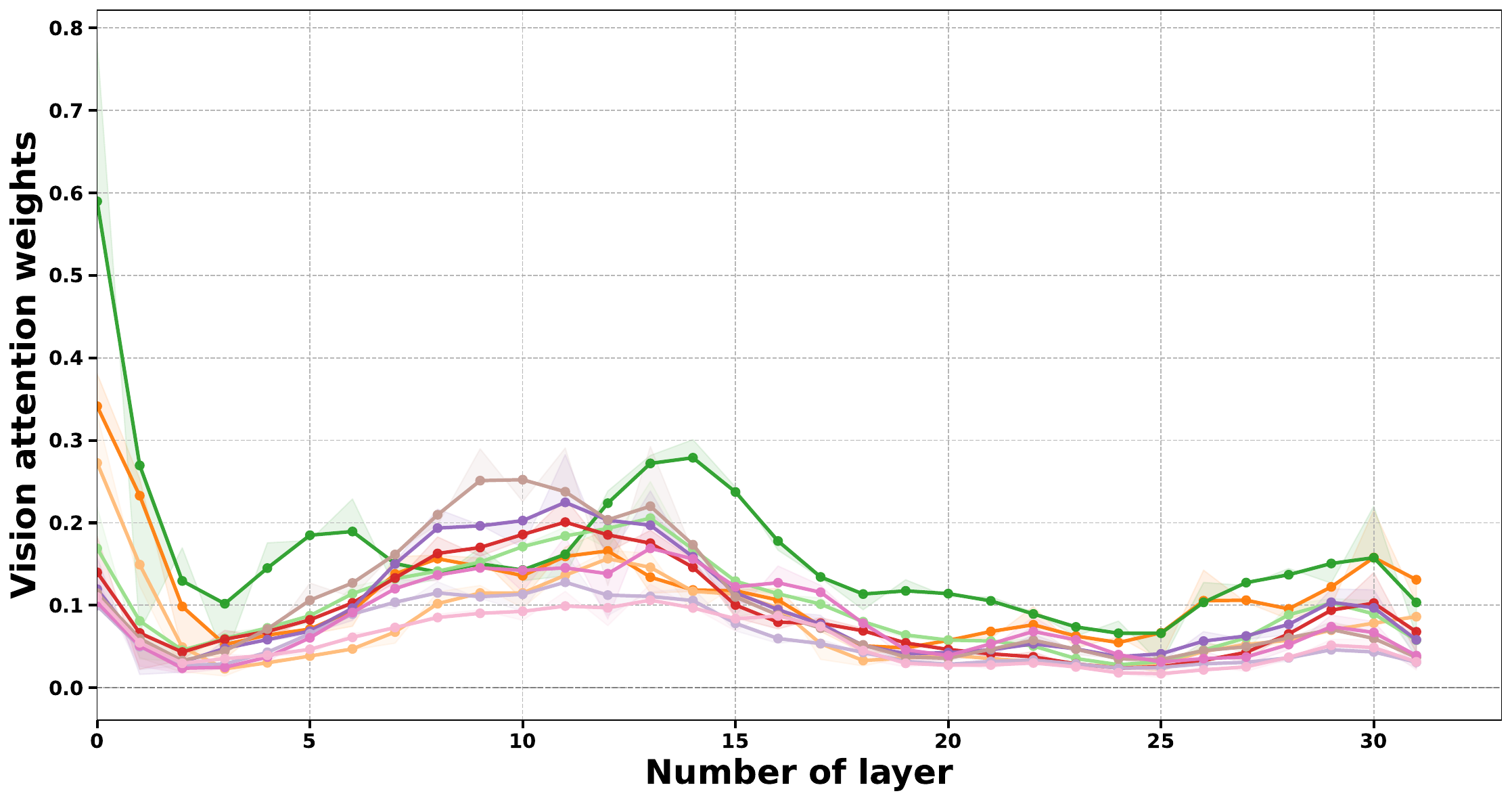}
      \caption{ Layer-wise visual attention of Idefics3-8B. }
      \label{fig:prior-8b}
    \end{figure}

\begin{figure}[tbp]
    \centering
      \includegraphics[width=0.45\textwidth]{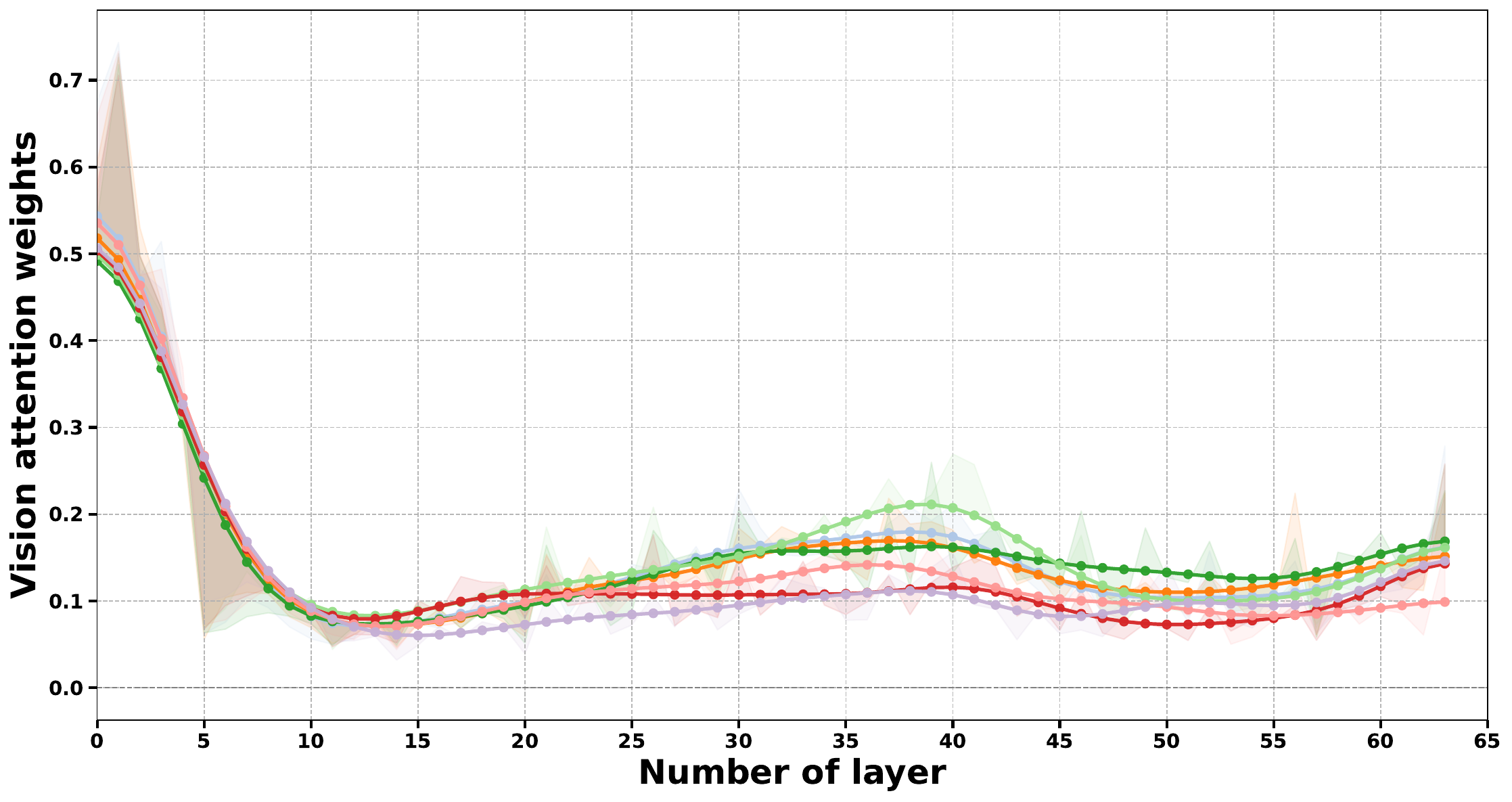}
      \caption{Layer-wise visual attention of InternVL2.5-38B. }
      \label{fig:prior-38b}
    \end{figure}

    \paragraph{Layer-Wise Attention Patterns}
    Figures~\ref{fig:prior-8b} and~\ref{fig:prior-38b} plot the average per-layer attention weights that each model's decoder assigns to visual tokens when processing vision inputs. In both the 8B and 38B variants, we observe a clear decay: the first few layers exhibit high visual attention, and then steadily decline toward deeper layers. This consistent decay profile confirms our multimodal prior hypothesis across both scales, justifying the use of an exponential decay schedule guided by these attention statistics.

\begin{figure}[tbp]
    \centering
      \includegraphics[width=0.45\textwidth]{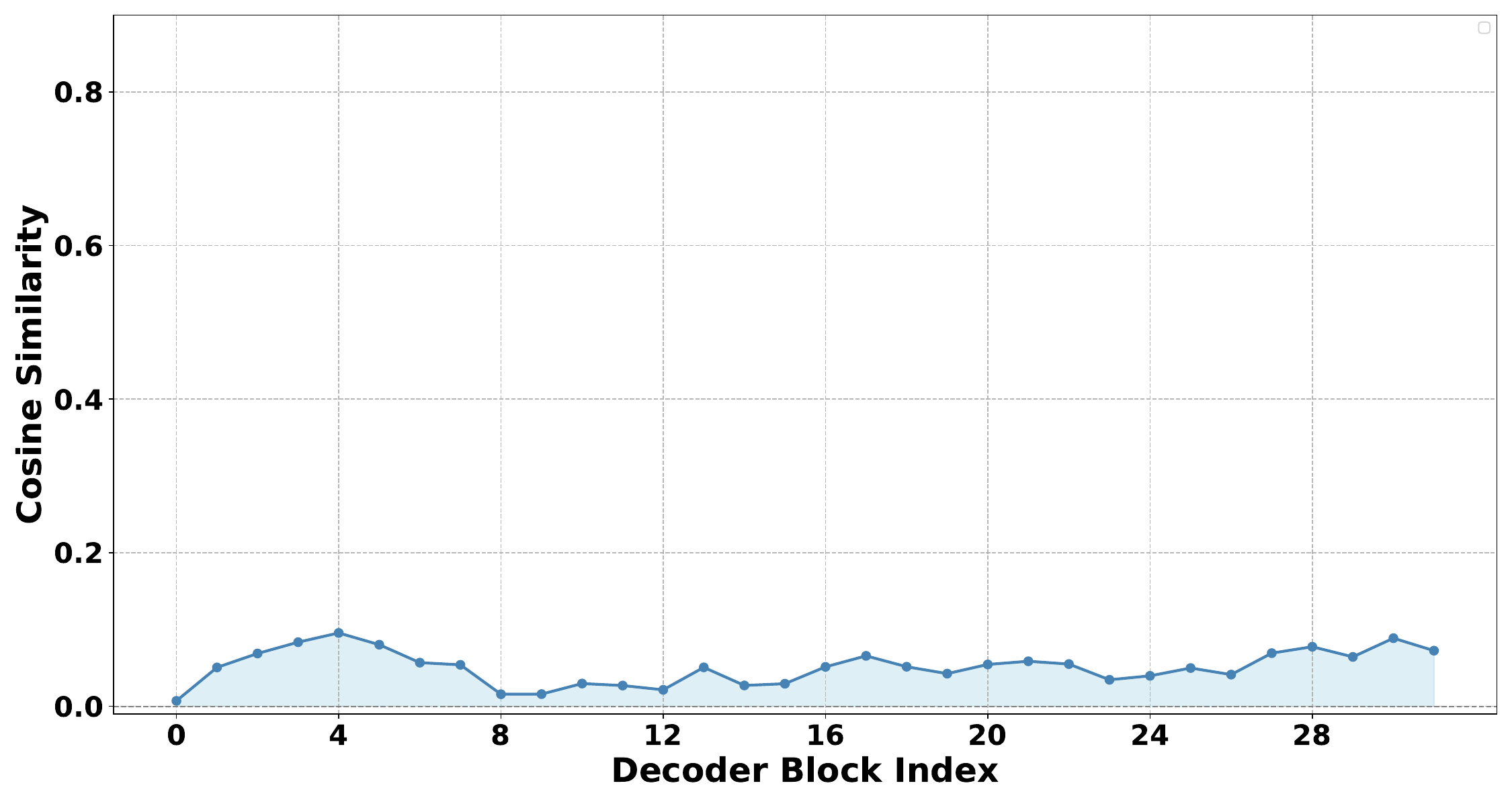}
      \caption{Cosine similarity between task vectors of vision-finetuned (Idefics3-8B) and reasoning-finetuned (DeepSeekDistil-LLaMA3-8B) models at each decoder block.}
      \label{fig:cosine-8b}
    \end{figure}

\begin{figure}[tbp]
    \centering
      \includegraphics[width=0.45\textwidth]{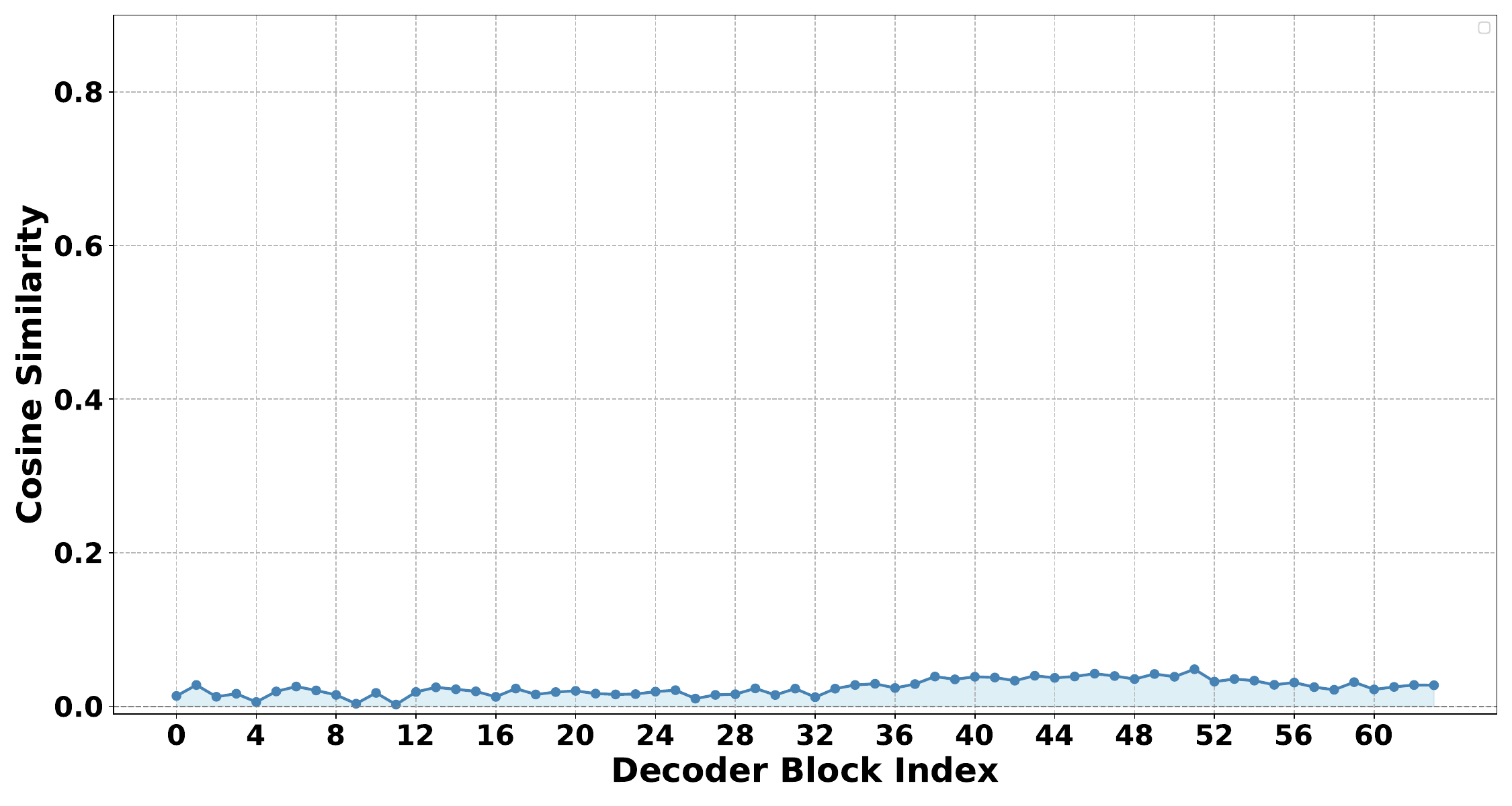}
      \caption{Cosine similarity between task vectors of vision-finetuned (InternVL2.5-38B) and reasoning-finetuned (QwQ-32B) models at each decoder block.}
      \label{fig:cosine-38b}
    \end{figure}

    \paragraph{Task-Vector Cosine Similarity}
    Figures~\ref{fig:cosine-8b} and~\ref{fig:cosine-38b} report the cosine similarity between the task vectors extracted from vision-finetuned models (Idefics3-8B / InternVL2.5-38B) and those from reasoning-finetuned counterparts (DeepSeekDistil-LLaMA3-8B / QwQ-32B) at each decoder block. In both cases, the similarity values remain close to zero across almost all layers, confirming that the vision and reasoning task vectors occupy nearly orthogonal subspaces. This orthogonality underpins our linear fusion derivation via Taylor expansion, ensuring that the combined update remains a meaningful superposition of the two modalities without destructive interference.

\subsection{Proof of NTK Linearization (Property)}
\label{app:proof-NTK}
Under the Neural Tangent Kernel (NTK) regime for wide transformers, the network behavior near initialization $\theta_0$ admits:
\begin{equation}
f(x;\theta_0 + \Delta\theta) = f(x;\theta_0) + \nabla_\theta f(x;\theta_0)^\top\Delta\theta + \mathcal{O}(\|\Delta\theta\|^2/\sqrt{\text{width}}). 
\end{equation}
For task-specific loss $\mathcal{L}_t^{(l)}$, the Hessian exhibits two key properties:
\textbf{Jacobian dominance}: Residual terms vanish due to $\nabla^2 f^{(l)} \sim \mathcal{O}(1/\sqrt{\text{width}})$. 
\textbf{Isotropy}: Gradient directions become nearly orthogonal in high dimensions. 
This leads to the layer-wise Hessian approximation:
\begin{equation}
\nabla^2\mathcal{L}_t^{(l)}(\theta) \approx \mathbb{E}_x\left[\nabla f^{(l)}\nabla f^{(l)\top}\right] \approx \delta_t^{(l)}I_{d_l}, 
\end{equation}
where the curvature scalar:
\begin{equation}
\delta_t^{(l)} = \frac{1}{d_l}\mathrm{Tr}\left(\nabla^2\mathcal{L}_t^{(l)}(\theta_0^{(l)})\right) = \frac{1}{d_l}\mathbb{E}_x\left[\|\nabla f^{(l)}(x;\theta_0^{(l)})\|^2\right], 
\end{equation}
captures the average gradient magnitude at initialization.

\subsection{Proof of Task-Vector Orthogonality (Property)}
\label{app:proof-orthogonality}
Figure~\ref{fig:prior} shows that, vision-fine-tuning and reasoning-fine-tuning update disjoint aspects of the LLM's representations. 
Concretely, let
\begin{equation}
  \tau_V^{(l)} = \theta_V^{(l)} - \theta_0^{(l)},
  \quad
  \tau_R^{(l)} = \theta_R^{(l)} - \theta_0^{(l)}.
\end{equation}
We observe that
\begin{equation}
  (\tau_V^{(l)})^\top \tau_R^{(l)} = \mathcal{O}(\varepsilon),
\end{equation}
where $\varepsilon$ is a small constant, indicating near-orthogonality.

\subsection{Proof of Taylor Expansion of Layer Task Loss Difference (Lemma)}
\label{app:proof-ltld}

To quantify the impact of parameter fusion on task performance, we analyze the Layer-wise Task Loss Difference (LTLD) through a rigorous Taylor expansion approach. 
The fused parameters combine both task updates through convex combination:
\begin{equation}
    \theta_{f}^{(l)} = \theta_0^{(l)} + \sum_{t \in \{V, R\}} \lambda_t^{(l)}\tau_t^{(l)}, \quad \text{where } \lambda_t^{(l)} \in [0,1], 
\end{equation}
where, $\theta_0^{(l)}$ denotes the pre-trained initialization at layer $l$, 
$\theta_t^{(l)} = \theta_0^{(l)} + \tau_t^{(l)}$ denotes the fine-tuned parameters for task $t \in \{V, R\}$, 
$\tau_t^{(l)}$ denotes the task-vector update from $\theta_0^{(l)}$ to $\theta_t^{(l)}$. 

\textbf{Fusion Residual Vector}: Measures deviation from optimal task parameters:
    \begin{equation}
        h_t^{(l)} = \theta_{f}^{(l)} - \theta_t^{(l)} = \sum_{k\neq t \in \{V, R\}} \lambda_k^{(l)}\tau_k^{(l)} - (1-\lambda_t^{(l)})\tau_t^{(l)}. 
    \end{equation}
\textbf{Interpolation Path}: 
Defines a linear trajectory in parameter space from the fine-tuned parameters $\theta_t^{(l)}$ to the fused parameters $\theta_f^{(l)}$, enabling exact Taylor expansion along the fusion direction:
\begin{equation}
    \gamma_t^{(l)}(\beta) = \theta_t^{(l)} + \beta h_t^{(l)}, \quad \beta \in [0,1].
\end{equation}
\noindent\textbf{Taylor Expansion Analysis:} Applying second-order expansion along $\gamma_t^{(l)}$:
\begin{equation}
    \begin{aligned}
    \mathcal{L}_t^{(l)}(\theta_{f}^{(l)},x_t) &= \mathcal{L}_t^{(l)}(\theta_t^{(l)} + h_t^{(l)},x_t) \\
    &= \mathcal{L}_t^{(l)}(\theta_t^{(l)},x_t) + \nabla\mathcal{L}_t^{(l)}(\theta_t^{(l)},x_t)^\top h_t^{(l)} \\
    &\quad + \frac{1}{2}h_t^{(l)\top}\left(\int_0^1\nabla^2\mathcal{L}_t^{(l)}(\gamma_t^{(l)}(\beta))d\beta\right)h_t^{(l)}
    \end{aligned}
\end{equation}
Under fine-tuning convergence $\nabla\mathcal{L}_t^{(l)}(\theta_t^{(l)},x_t) \approx 0$, we obtain:
\begin{equation}
\begin{aligned}
\text{LTLD}_t^{(l)} &= \mathcal{L}_t^{(l)}(\theta_{f}^{(l)},x_t) - \mathcal{L}_t^{(l)}(\theta_t^{(l)},x_t) \\
                &= \frac{1}{2}h_t^{(l)\top}\left(\int_0^1\nabla^2\mathcal{L}_t^{(l)}(\gamma_t^{(l)}(\beta))d\beta\right)h_t^{(l)}. 
\end{aligned}
\end{equation}
This derivation establishes the theoretical foundation for our layer-wise fusion analysis, connecting parameter perturbations to task performance through differentiable geometry.
    
\subsection{Derivation of Closed-Form Fusion Weights}
\label{app:proof-weights}
To enable efficient model merging while preserving task performance, we derive theoretically-grounded fusion weights through layer-wise NTK and Task-Vector Orthogonality analysis. 
The key insight is that LLMs exhibit approximately quadratic loss landscapes under NTK, permitting closed-form solutions. 
Under the NTK linearization regime for LLMs, we first analyze the layer-wise behavior. 
For any layer $l$ with initialization $\theta_0^{(l)}$ and perturbation $\tau^{(l)} = \theta^{(l)} - \theta_0^{(l)}$, the parameter admits the first-order approximation:
\begin{equation}
f^{(l)}(x;\theta^{(l)}) \approx f^{(l)}(x;\theta_0^{(l)}) + \underbrace{\nabla_{\theta^{(l)}}f^{(l)}(x;\theta_0^{(l)})^\top\tau^{(l)}}_{\text{Linear term}} + \mathcal{O}(\|\tau^{(l)}\|^2)
\end{equation}
This linearity emerges in wide transformers where the network's output changes linearly with parameter perturbations.

For the quadratic loss $\mathcal{L}_t^{(l)}(\theta^{(l)},x_t) = \frac{1}{2}\|f^{(l)}(x_t;\theta^{(l)}) - y_t\|^2$, we compute the Hessian:
\begin{align}
\nabla^2\mathcal{L}_t^{(l)} &= \nabla f^{(l)}(x_t;\theta^{(l)}) \nabla f^{(l)}(x_t;\theta^{(l)})^\top \notag \\
&\quad + \underbrace{(f^{(l)}(x_t;\theta^{(l)}) - y_t)^\top \nabla^2 f^{(l)}(x_t;\theta^{(l)})}_{\text{Vanishes under NTK regime}}
\end{align}
Under the NTK regime, the second term becomes negligible due to two intrinsic properties of wide neural networks: 1) the output residual $\|f^{(l)}-y_t\|$ vanishes with near-optimal fine-tuning (guaranteed by NTK's convex-like optimization landscape), and 2) the Hessian $\nabla^2 f^{(l)}$ shrinks as $\mathcal{O}(1/\sqrt{\text{width}})$ (a direct consequence of NTK's linearization effect). 

Under NTK conditions:
\begin{equation}
\nabla^2\mathcal{L}_t^{(l)}(\theta^{(l)}) \approx \nabla f^{(l)}(x_t;\theta_0^{(l)}) \nabla f^{(l)}(x_t;\theta_0^{(l)})^\top
\end{equation}
Taking the isotropic approximation for layers:
\begin{equation}
\nabla^2\mathcal{L}_t^{(l)} \approx \delta_t^{(l)}I_{d_l}, \quad \delta_t^{(l)} = \frac{1}{d_l}\text{tr}\left(\nabla f^{(l)}(x_t;\theta_0^{(l)}) \nabla f^{(l)}(x_t;\theta_0^{(l)})^\top\right)
\end{equation}
where $d_l$ is the parameter dimension at layer $l$. 
This follows from the observation that gradient directions become nearly orthogonal in high dimensions.

Substituting the Hessian approximation into the Taylor remainder:
\begin{equation}
  \int_0^1 \nabla^2 \mathcal{L}_t^{(l)}(\gamma_t^{(l)}(\beta)) \,d\beta
  \approx \delta_t^{(l)} I, 
\end{equation}
collapses the quadratic form to a scalar multiple of $\|h_t^{(l)}\|^2$, yielding a tractable bound and closed-form solution for fusion weights.
The second-order Taylor remainder is bounded by the extremal eigenvalues of the integral Hessian. 
Substituting our isotropic approximation: 
\begin{align}
2\text{LTLD}_t^{(l)} &\leq h_t^{(l)\top}\left(\delta_t^{(l)}I_{d_l}\right)h_t^{(l)} \notag \\
&= \delta_t^{(l)}\|h_t^{(l)}\|^2 \notag \\
&= \delta_t^{(l)}\left\|\sum_{k\neq t \in \{V, R\}}\lambda_k^{(l)}\tau_k^{(l)} - (1-\lambda_t^{(l)})\tau_t^{(l)}\right\|^2
\end{align}
where we used the identity $\|a + b\|^2 = \|a\|^2 + \|b\|^2 + 2a^\top b$. 
The cross-term vanishes due to task vector orthogonality. 
This demonstrates that the layer-wise task loss difference is governed by the product of the initialization curvature $\delta_t^{(l)}$ and the squared fusion residual norm $\|h_t^{(l)}\|^2$, justifying our layer-wise analysis.

Next, we show the optimal layer-wise fusion weights. 
The layer-wise analysis enables independent optimization at each layer. 
The LALD decomposes into layer-specific terms:
\begin{equation}
\begin{aligned}
    &\text{LALD}^{(l)} \leq \\
    &\sum_{t \in \{V, R\}}\frac{\delta_t^{(l)}}{2}\|\tau_t^{(l)}\|^2\left[(1-\lambda_t^{(l)})^2\|\tau_t^{(l)}\|^2 + \sum_{k\neq t \in \{V, R\}}(\lambda_k^{(l)})^2\|\tau_k^{(l)}\|^2\right]
\end{aligned}
\end{equation}
This additive structure permits layer-wise optimization by solving $L$ (where $L$ is the total number of layers) independent problems:
\begin{equation}
\{\lambda_t^{(l)}\}_{t \in \{V, R\}} = \argmin J^{(l)}, \quad \forall l = 1,...,L
\end{equation}

Focusing on a single layer (omitting superscript $(l)$ for clarity), we reformulate the objective:
\begin{equation}
\begin{aligned}
J &= \sum_{t \in \{V, R\}} \frac{\delta_t}{2}\|\tau_t\|^2\Big[(1-\lambda_t)^2\|\tau_t\|^2 + \sum_{k\neq t  \in \{V, R\}}\lambda_k^2\|\tau_k\|^2\Big] \\
&= \frac{1}{2}\sum_{t \in \{V, R\}} \delta_t\|\tau_t\|^4(1-\lambda_t)^2 + \frac{1}{2}\sum_{t \in \{V, R\}}\sum_{k\neq t  \in \{V, R\}}\delta_t\lambda_k^2\|\tau_t\|^2\|\tau_k\|^2
\end{aligned}
\end{equation}

Under task orthogonality $\tau_V^\top\tau_R = 0$, the cross-derivative terms vanish, simplifying the gradient to: 
\begin{equation}
\begin{aligned}
\frac{\partial J}{\partial\lambda_t} &= -\delta_t\|\tau_t\|^4(1-\lambda_t) + \lambda_t\|\tau_t\|^2\sum_{k\neq t  \in \{V, R\}}\delta_k\|\tau_k\|^2 \\
0 &= -\delta_t\|\tau_t\|^4 + \lambda_t\left[\delta_t\|\tau_t\|^4 + \|\tau_t\|^2\sum_{k\neq t  \in \{V, R\}}\delta_k\|\tau_k\|^2\right]
\end{aligned}
\end{equation}

Solving the linear system yields:
\begin{equation}
\lambda_t = \frac{\delta_t\|\tau_t\|^4}{\delta_t\|\tau_t\|^4 + \|\tau_t\|^2\sum_{k\neq t}\delta_k\|\tau_k\|^2} = \frac{\delta_t\|\tau_t\|^2}{\sum_{k \in \{V, R\}}\delta_k\|\tau_k\|^2}
\end{equation}

Under the uniform curvature assumption $\delta_k \approx \delta_0$ (arising from NTK's layer-wise gradient statistics in wide networks, where $\frac{1}{d_l}\mathrm{Tr}(\nabla f^{(l)}\nabla f^{(l)\top})$ becomes task-invariant as $d_l \to \infty$), the solution simplifies to:
\begin{equation}
    \lambda^{(l)}_t = \frac{\|\tau^{(l)}_t\|^2}{\sum_{k  \in \{V, R\}}\|\tau^{(l)}_k\|^2} \quad \Rightarrow \quad \lambda^{(l)}_t \propto \|\tau^{(l)}_t\|^2
\end{equation}

This closed-form solution adaptively suppresses interfering task vectors while preserving target task information at each layer.

  \subsection{Derivation of Attention-Guided Exponential Decay Priors}
  \label{app:attention_priors}
  In Section~\ref{sec:prior} we introduce an attention-guided exponential schedule for modality priors. 
  Here, we provide a step-by-step derivation.
  
  \paragraph{Modeling the decay.}
  We observe the layer-wise visual-attention ratio
  \begin{equation}
    a_l \;=\; \frac{1}{N}\sum_{x,\,h}\frac{\mathrm{Attn}_{l\to\text{vis}}(x,h)}{\mathrm{Attn}_{l\to\text{vis}}(x,h) + \mathrm{Attn}_{l\to\text{text}}(x,h)}
    \quad \forall\,l=1,\dots,L,
  \end{equation}
  where $x$ indexes inputs, $h$ indexes attention heads, and $N$ is a normalization factor.  We posit an exponential decay model:
  \begin{equation}\label{eq:exp_model}
    a_l \approx C\,e^{-\alpha\,l},
  \end{equation}
  where $C>0$ and $\alpha>0$ are unknown constants.
  
  \paragraph{Logarithmic linearization.}
  Taking natural logarithm on both sides of Eq. (~\ref{eq:exp_model}), we obtain
  \begin{equation}\label{eq:log_linear}
    \ln a_l \approx \ln C \; - \; \alpha\,l.
  \end{equation}
  Define
  \(y_l = \ln a_l\),
  \(x_l = l\),
  \(b_0 = \ln C\),
  \(b_1 = -\alpha\).
  Then Eq. (~\ref{eq:log_linear}) becomes a linear regression problem:
  \begin{equation}
    y_l \approx b_0 + b_1\,x_l.
  \end{equation}
  
  \paragraph{Least-squares estimation.}
  We collect the dataset $\{(x_l,y_l)\}_{l=1}^L$ and solve the normal equations for the least-squares fit:
  \begin{equation}
  \begin{aligned}
    b_1 &= \frac{\sum_{l=1}^L (x_l-\bar x)(y_l-\bar y)}{\sum_{l=1}^L (x_l-\bar x)^2},\\
    b_0 &= \bar y - b_1\,\bar x,
  \end{aligned}
\end{equation}
  where $\bar x = \tfrac1L\sum_{l}x_l$ and $\bar y = \tfrac1L\sum_{l}y_l$. 
  We then recover
  \begin{equation}
    \hat\alpha = -\,b_1,
    \quad
    \hat C = e^{\,b_0}.
  \end{equation}
  
  \paragraph{Constructing modality priors.}
  Finally, we define the normalized exponential priors
  \begin{equation}
    w_V^{(l)} = \frac{e^{-\hat\alpha\,l}}{\sum_{j=1}^L e^{-\hat\alpha\,j}},
    \quad
    w_R^{(l)} = 1 - w_V^{(l)}.
  \end{equation}
  These priors smoothly interpolate from strong visual emphasis in early layers to strong reasoning emphasis in late layers, and can be replaced by alternative data-driven schedules if desired.

\end{document}